\documentclass[runningheads]{llncs}

\usepackage{eccv}
\usepackage{eccvabbrv}
\usepackage{graphicx}
\usepackage{booktabs}
\usepackage[accsupp]{axessibility}  
\usepackage[pagebackref,breaklinks,colorlinks,citecolor=eccvblue]{hyperref}
\usepackage{orcidlink}

\usepackage[dvipsnames]{xcolor}

\usepackage{graphicx}
\usepackage{cuted}
\usepackage{capt-of}
\usepackage{amsmath}
\usepackage{caption}
\usepackage{subcaption}
\usepackage{amssymb}
\usepackage{multibib}
\usepackage{wrapfig}
\usepackage{multirow}
\usepackage{multicol}
\usepackage{booktabs}
\usepackage{enumitem}
\usepackage{pifont}
\usepackage{xcolor}
\usepackage{lipsum}
\usepackage{float}

\newcommand{\cmark}{\textcolor{Green}{\ding{51}}}
\newcommand{\xmark}{\textcolor{Red}{\ding{55}}}

\newcommand{\cutsectionup}{\vspace*{-0.15in}}

\newcommand{\cutparagraphup}{\vspace*{-0.15in}}

\renewcommand{\paragraph}[1]{{\parindent0pt \textbf{#1.}}}

\newcommand{\modelname}{Chameleon}

\begin{document}

\title{\modelname{}: A Data-Efficient Generalist for Dense Visual Prediction in the Wild}

\titlerunning{A Data-Efficient Generalist for Dense Visual Prediction in the Wild}

\author{
  \footnotesize{Donggyun Kim$^1$, Seongwoong Cho$^1$, Semin Kim$^1$, Chong Luo$^2$, Seunghoon Hong$^1$}
  \vspace{-0.2cm}
}

\authorrunning{D. Kim et al.}

\institute{$^1$School of Computing, KAIST
\qquad
$^2$Microsoft Research Asia
}

\maketitle

{\centering
\includegraphics[width=\textwidth]{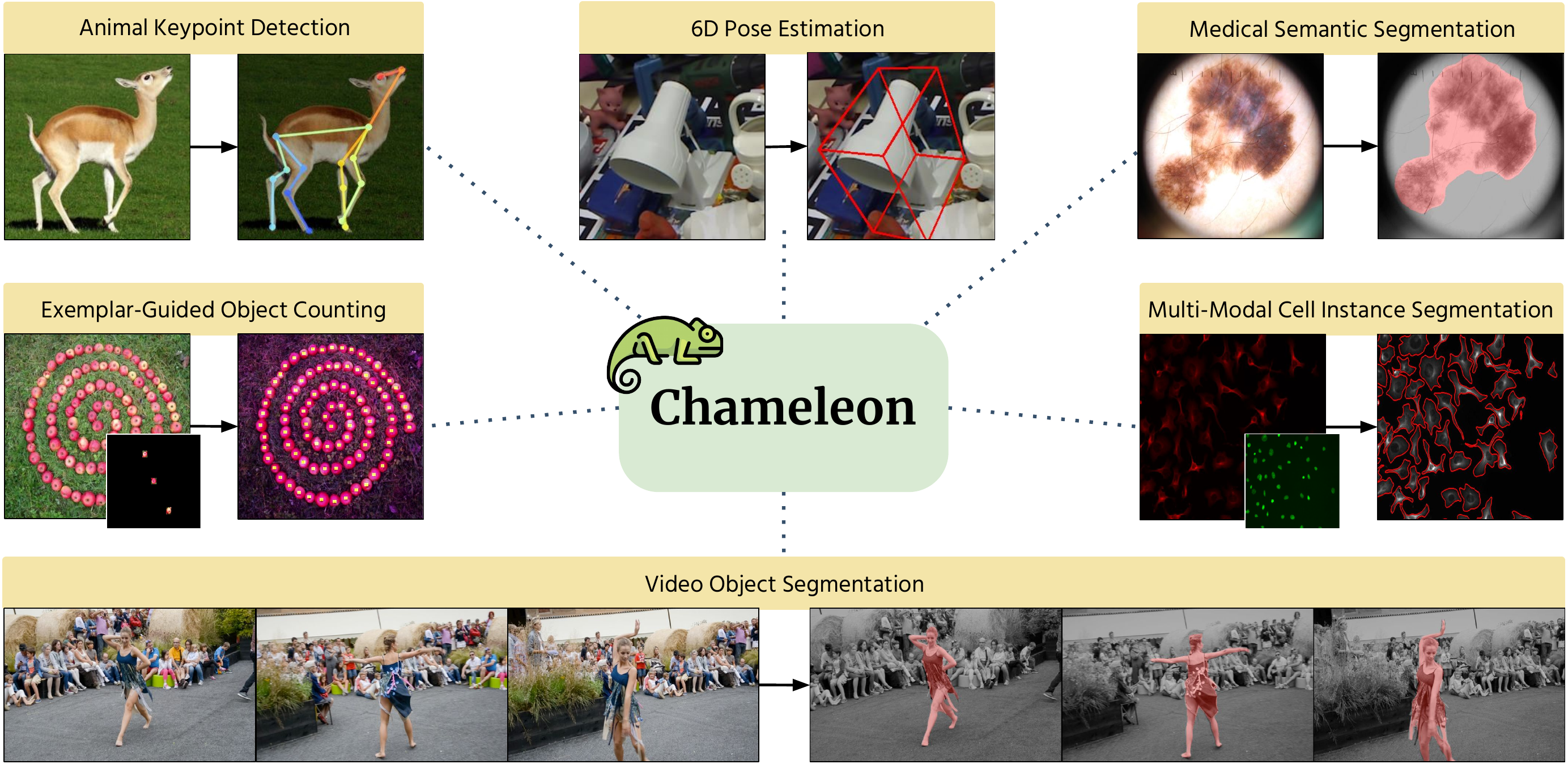}
\captionof{figure}{
\modelname{} is a data-efficient generalist that can adapt to various \textbf{unseen} dense visual prediction tasks in the wild with arbitrary output structures using a handful of examples (dozens). It can also learn to utilize multi-modal inputs and user-interactions. 
\label{fig:result-preview}}}

\vspace{-0.6cm}
\begin{abstract}
Large language models have evolved data-efficient generalists, benefiting from the universal language interface and large-scale pre-training.
However, constructing a data-efficient generalist for dense visual prediction presents a distinct challenge due to the variation in label structures across different tasks.
Consequently, generalization to unseen dense prediction tasks in the low-data regime is not straightforward and has received less attention from previous vision generalists.
In this study, we explore a universal model that can flexibly adapt to unseen dense label structures with a few examples, enabling it to serve as a data-efficient vision generalist in diverse real-world scenarios.
To this end, we base our method on a powerful meta-learning framework and explore several axes to improve its performance and versatility for real-world problems, such as flexible adaptation mechanisms and scalability.
We evaluate our model across a spectrum of unseen real-world scenarios where low-shot learning is desirable, including video, 3D, medical, biological, and user-interactive tasks.
Equipped with a generic architecture and an effective adaptation mechanism, our model flexibly adapts to all of these tasks with at most 50 labeled images, showcasing a significant advancement over existing data-efficient generalist approaches.
Codes are available at {\footnotesize\url{https://github.com/GitGyun/chameleon}}.
\vspace{-0.3cm}
\end{abstract}

\section{Introduction}
\vspace{-0.3cm}
\label{sec:intro}
Generalist models have gained significant attention across various fields~\cite{brown2020language,li2023uni,reed2022a,ibarz2022generalist,alayrac2022flamingo} with their data efficiency in learning new tasks.
In contrast to specialist models designed specifically to achieve certain tasks, generalist models aim to address a broad range of tasks, including those unseen during training.
Moreover, generalist models have even begun competing with specialist models while using much less supervision attributed to incorporating two key ingredients: (1) a universal learning framework and (2) large-scale pre-training.
For instance, large language models~\cite{brown2020language,ouyang2022training,openai2023gpt} have exhibited exceptional generalization abilities, benefiting from the universal nature of natural language and unsupervised pre-training on extensive corpora.
Similarly, in the fields of algorithmic learning and reinforcement learning, large-scale training through universal interfaces—graph neural networks~\cite{ibarz2022generalist,mahdavi2022towards,rodionov2024neural} and transformers~\cite{reed2022a,shridhar2023perceiver,schubert2023generalist}, respectively—have demonstrated decent generalization performance.

\begin{figure}[t!]
    \centering
    \includegraphics[width=\linewidth]{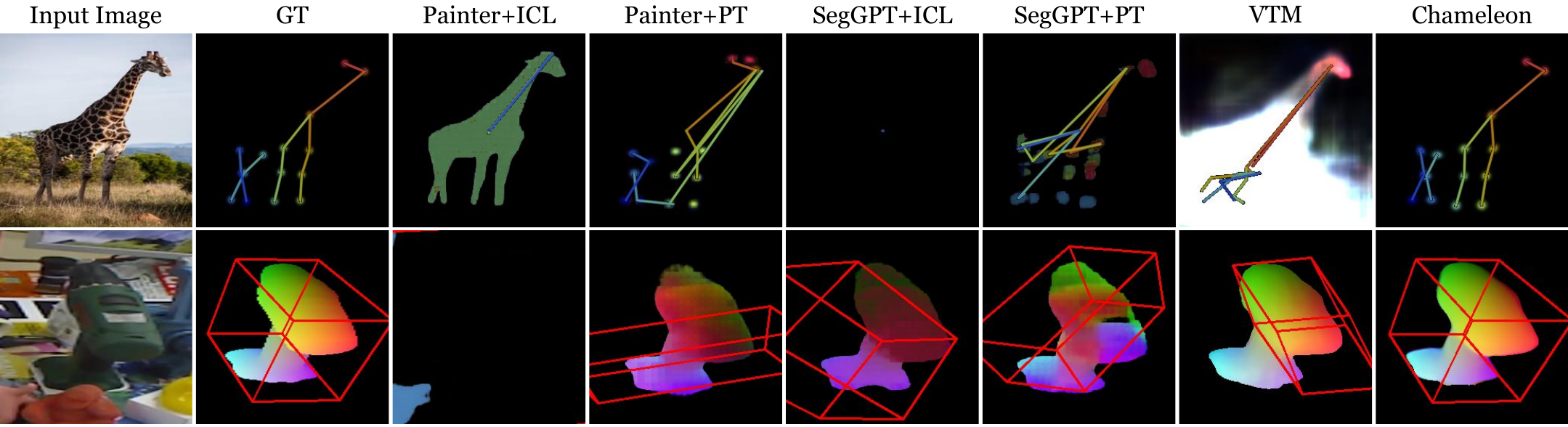}
    \vspace{-0.6cm}
    \caption{
    Existing generalist models struggles to learn out-of-distribution tasks of unseen label semantics (6D pose) or structure (animal keypoint) during training.
    ICL and PT denote in-context learning and prompt tuning is used for adaptation, respectively.
    }
    \vspace{-0.6cm}
    \label{fig:oodgen_comparison}
\end{figure}

However, building a data-efficient generalist for dense visual prediction tasks, which involve high-dimensional outputs with vastly diverse structure and semantics~\cite{awais2023foundational}, remains less explored.
Most of the prior efforts for general dense visual prediction~\cite{lu2023unifiedio,kolesnikov2022uvim,chen2023generalist} mainly focus on unifying a range of \emph{pre-defined} tasks into a single model, rather than generalizing to \emph{unseen} tasks.
Conversely, in-context learning approaches~\cite{wang2022images,wang2023seggpt} attempt to solve various tasks with few demonstrations by framing the dense prediction as an image-to-image translation problem.
Yet, these methods often struggle to generalize to out-of-distribution tasks that have distinct output structures and semantics unseen during training, which limits their applicability to various real-world problems.
Figure~\ref{fig:oodgen_comparison} highlights the necessity of a more flexible adaptation mechanism in building data-efficient vision generalists for arbitrary dense visual prediction.

In this work, we aim to explore the potential of a powerful and flexible data-efficient generalist for diverse real-world dense prediction tasks.
To this end, we build our method based on the framework of Visual Token Matching (VTM)~\cite{kim2023universal}, which directly focuses on out-of-distribution generalization in low-data regimes.
First, we design an encoding mechanism to incorporate varying numbers and types of input modalities, which expands the scope of adaptable tasks and addresses drifts in data modality or multi-input scenarios.
Second, we enhance the task-specific adaptation mechanism by introducing a task-adaptive feature re-weighting module in the hierarchical architecture.
Lastly, we enlarge and diversify the meta-training data to make the model acquire more general prior knowledge, as well as scale up the modal capacity and resolution.
We meta-train the model on a large-scale dataset constructed by combining six existing datasets from diverse domains, which consists of 17 different dense visual prediction tasks.

We evaluate our method, termed \modelname, in six downstream benchmarks composed of unique and unseen structured outputs, including tasks with video, 3D, medical and biological data, and user-interactive tasks.
Our results show that existing in-context learning approaches, even if they are empowered by prompt tuning at test time, have limited generalization capability to out-of-distribution tasks, while our method successfully adapts to each scenario using at most 50 labeled examples per task, significantly outperforming the generalist baselines.
Our extensive analyses also suggest that effective encoding mechanism with flexible adaptation and meta-training on a rich dataset are the key factors of successful generalization to out-of-distribution tasks.

\vspace{-0.3cm}
\section{Related Work}
\vspace{-0.3cm}

\paragraph{Generalist Models}
Recently, generalist models have emerged as an effective approach to tackle a variety of tasks seamlessly within a single framework.
In computer vision, generalist models for dense visual prediction have mainly focused on multi-task learning and prompting approaches.
Multi-task learning approaches~\cite{lu2023unifiedio,kolesnikov2022uvim,chen2023generalist,ye2022taskprompter,geng2023instructdiffusion} train a unified architecture to solve diverse tasks, but they require a large amount of labeled data for each task and lack generalization ability to unseen tasks.
In-context learning approaches~\cite{wang2023seggpt,wang2022images} have proposed to address unseen tasks but they either address in-distribution tasks whose label structures or semantics are seen during training or focus on segmentation tasks.

\paragraph{Few-shot Learning}
Few-shot learning also targets a wide range of tasks within a single framework, but its main focus is on learning from a few labeled examples.
In computer vision, most attention is paid to a specific set of tasks with dedicated architectures, such as image classification~\cite{vinyals2016matching,snell2017prototypical,liu2020universal,bateni2020improved}, object detection~\cite{fan2020few, wang2020frustratingly,han2022few}, and semantic segmentation~\cite{min2021hypercorrelation,hong2022cost,shaban2017one}, which are not suitable for out-of-distribution generalization.
Visual Token Matching~\cite{kim2023universal} proposes a universal few-shot learning problem for dense visual prediction, whose main focus is out-of-distribution generalization to arbitrary tasks with only a few labels.
However, it has only been demonstrated in a constrained setting where both the meta-training and testing are from the same narrow domains (\emph{i.e.}, indoor scene), leaving its potential as a generalist in various real-world applications in question.
\vspace{-0.1cm}
\cutsectionup
\section{Approach}
\vspace{-0.2cm}
\label{sec:approach}

Chameleon is a data-efficient generalist based on the Visual Token Matching~\cite{kim2023universal} framework, improving its design and scalability to address low-shot learning problems in broader and more challenging real-world applications. 
In this section, we first present our problem setting and overall framework, then describe our improved encoder designs for handling variable multi-modal inputs (Section~\ref{subsec:multi_input}) and enhancing the adaptation mechanism (Section~\ref{subsec:feature_reweighting}).

\paragraph{Problem Setting}
\modelname{} is designed as a versatile model capable of learning arbitrary dense prediction tasks with minimal labeled data.
Formally, given a (multi-modal) query image $X^q\in\mathbb{R}^{3I_\mathcal{T} \times H_\mathcal{T} \times W_\mathcal{T}}$, our goal is to produce the per-pixel label $Y^q\in\mathbb{R}^{O_\mathcal{T} \times H_\mathcal{T} \times W_\mathcal{T}}$ of an arbitrary task $\mathcal{T}$ adaptively based on the small number of labeled examples  $\mathcal{S}_\mathcal{T}$ (\emph{i.e.} support set) by:
\begin{equation}
    Y^q = \mathcal{F}(X^q; \mathcal{S}_\mathcal{T}), \quad \mathcal{S}_\mathcal{T} = \{(X^i, Y^i)\}_{i \le N}.
    \label{eqn:few-shot}
\end{equation}
Importantly, \modelname{} does not presuppose specific priors on dense prediction tasks, allowing its application to various \emph{unseen} tasks with the unique amount of inputs $I_\mathcal{T}$ and output channels $O_\mathcal{T}$ as well as their semantics and spatial resolutions $(H_\mathcal{T}, W_\mathcal{T})$.
This makes it applicable to a wide range of real-world problems whose inputs and outputs are defined over pixels, such as segmentation, stereo depth estimation, dense pose estimation, and exemplar-guided object counting, to name a few.

\paragraph{Overall Framework}
To support versatility, \modelname{} employs the universal token matching framework~\cite{kim2023universal} that formulates dense prediction as a token-level matching problem between query and support images as follows:
\begin{equation}
    g\left(\mathbf{y}_k^q\right) = \sum_{i \le N} \sum_{j \le M} \sigma \left( f_\mathcal{T}\left(\mathbf{x}_k^q\right), f_\mathcal{T}\left(\mathbf{x}_j^i\right) \right) \cdot g\left(\mathbf{y}_j^i\right), \quad \forall~k \le M,
\label{eqn:matching}
\end{equation}
where $f_\mathcal{T}(\mathbf{x}_k)$ and $g(\mathbf{y}_k)$ denote the $k$-th token embeddings obtained by an image $X$ and a label $Y$, respectively, $\sigma$ is a similarity function, and $M$ is the number of tokens per image.
In this framework, the prediction for the $k$-th query token is produced by interpolating the support label embeddings based on its similarity to the support image embeddings.
To incorporate various similarities for dense prediction in a single framework, a small amount of task-specific parameters $\theta_\mathcal{T}$ are introduced in the image encoder to adapt the image token embeddings $f_\mathcal{T}(\mathbf{x}) = f(\mathbf{x}; \theta, \theta_\mathcal{T})$ while sharing the other parameters across all tasks.
After the matching in Eq.~(\ref{eqn:matching}) is performed, the predicted query token embeddings are decoded into the query label by a label decoder $h\approx g^{-1}$.

The training protocol consists of two stages: episodic meta-training and few-shot fine-tuning.
During episodic training, the whole model is trained with various dense prediction tasks sampled from a meta-training dataset to learn a general concept of matching.
At this stage, \modelname{} maintains and tunes separate sets of task-specific parameters $\theta_{\mathcal{T}}$ of the image encoder for each training task $\mathcal{T}_\text{train}$.
After meta-training, \modelname{} adapts to an unseen target task $\mathcal{T}_\text{test}$ by fine-tuning the task-specific parameters $\theta_\mathcal{T}$ with a small support set $\mathcal{S}_{\mathcal{T}_\text{test}}$.
To further adapt the model to unseen output structures, we also fine-tune a part of the label decoder $h$ (\emph{e.g.,} a linear head) while fixing the rest.

The key design of \modelname{} lies in how to produce the image token embeddings $f_\mathcal{T}(\mathbf{x})$.
Since the matching (Eq.~(\ref{eqn:matching})) requires the number of the tokens in the image and the label to be consistent, we need to design a flexible encoding mechanism that handles arbitrary input space with a varying number of modalities $I_\mathcal{T}$.
At the same time, the encoding mechanism should reflect the unique correlation between the input modalities, which varies significantly per task.
Another crucial component is the adaptation mechanism of the image encoder $f_\mathcal{T}(\mathbf{x}) = f(\mathbf{x}; \theta, \theta_\mathcal{T})$ and the choice of task-specific parameters $\theta_\mathcal{T}$.
It should be flexible enough to adapt in order to predict vastly diverse semantics and structures of labels that are unseen during training, while not overfitting to the small support set.
In the following sections, we explain how we design each component.

\cutparagraphup
\subsection{Encoder for Variable Input Images}
\vspace{-0.1cm}
\label{subsec:multi_input}

\begin{figure}[t!]
    \centering
    \includegraphics[width=\linewidth]{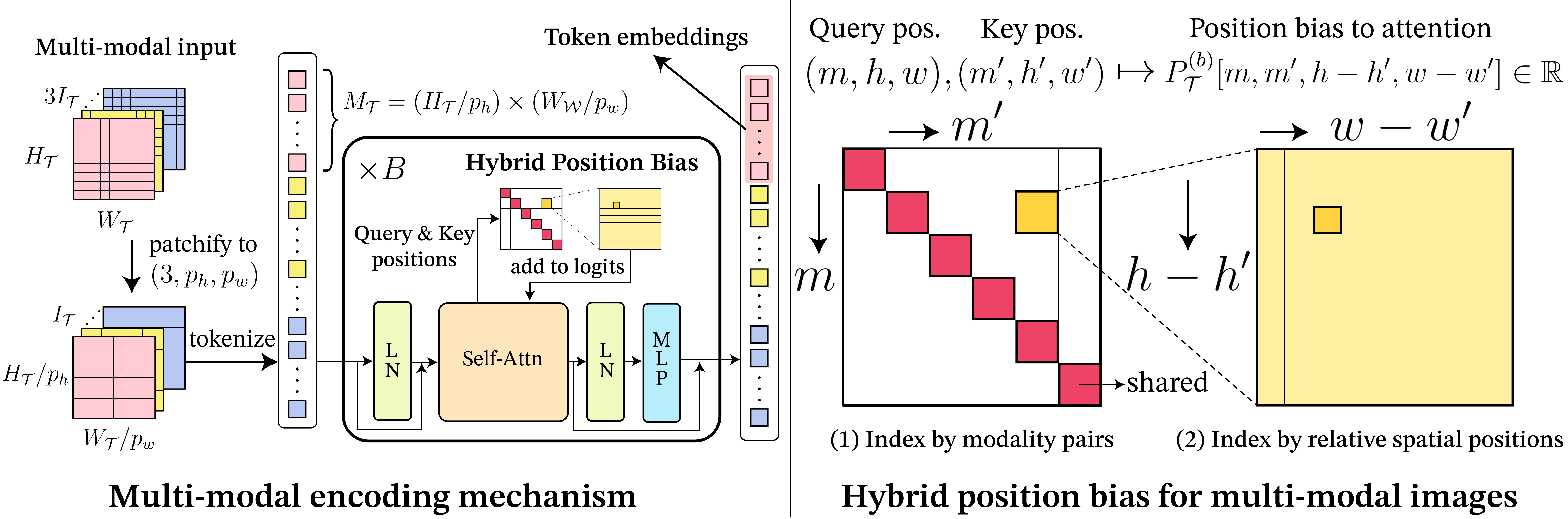}
    \vspace{-0.5cm}
    \caption{Encoding mechanism of the image encoder to handle multiple input images.}
    \label{fig:multi_input}
    \vspace{-0.5cm}
\end{figure}

To effectively handle tasks with varying numbers and types of input modalities, we design an encoding mechanism based on Transformer~\cite{vaswani2017attention} as illustrated in Figure~\ref{fig:multi_input}.
First, we patchify a multi-modal input $X \in \mathbb{R}^{3I_\mathcal{T} \times H_\mathcal{T} \times W_\mathcal{T}}$ with a fixed patch size $(3, p_h, p_w)$, which results in $I_\mathcal{T} \times M_\mathcal{T}$ tokens where $M_\mathcal{T} = (H_\mathcal{T} / p_h) \times (W_\mathcal{T} / p_w)$ denotes the number of tokens per modality.
Then we encode all the tokens at once by a transformer encoder, which contextualizes the token embeddings across modalities.
Importantly, we should also encode the positional information about tokens, such that the encoder can incorporate the varying relationship between input modalities as well as the spatial prior adaptively per task.
Besides the example-level contextualization, such information allows our model to learn and adapt global correlation across the input modalities per task. 

To model the positional relationships between the multi-modal tokens, we design a learnable positional embedding that extends the relative position bias~\cite{raffel2020exploring,bao2022beit}.
In each $b$-th attention layer, the position bias between a query token at position $(m, h, w)$ and a key token at position $(m^\prime, h^\prime, w^\prime)$ is computed by indexing a learnable embedding $P_\mathcal{T}^{(b)}$ as follows:
\begin{equation}
    P_\mathcal{T}^{(b)}[m, m^\prime, h - h^\prime, w - w^\prime] \in \mathbb{R}.
\end{equation}
The first two indices $(m, m^\prime)$ distinguish each modality pair, such that different types of \emph{inter-modal} interaction between tokens can be modeled.
Then the remaining indices $(h - h^\prime, w - w^\prime)$ distinguish the relative spatial positions, which effectively encodes the translation-equivariance along the spatial axes.
Note that we assign different embeddings $P_\mathcal{T}^{(b)}$ for each task as a part of task-specific parameters $\theta_\mathcal{T}$.
This ensures the encoder not only handles different numbers of positions but also adapts to contextualize distinct relationships between modalities of each task separately.
Having that the information from other modalities is contextualized to each modality, we use the first $M_\mathcal{T}$ tokens as image token embeddings for the matching (Eq.~(\ref{eqn:matching})).

\cutparagraphup
\subsection{Feature Modulation of the Image Encoder}
\vspace{-0.1cm}
\label{subsec:feature_reweighting}

To adapt to tasks with unseen semantics and structures of labels, \modelname{} modulates the image encoder in two ways.
First, the bias parameters $\mathbf{b}_\mathcal{T}$ of each image encoder layer are tuned separately for each task $\mathcal{T}$.
This has been proven to efficiently modulate the features in a transformer encoder~\cite{zaken2022bitfit,kim2023universal}.
Second, we introduce a feature re-weighting mechanism to associate different levels of image and label features.
While it is known that using multi-level image features is beneficial for dense prediction in general~\cite{lin2017feature,chen2017rethinking,ranftl2021vision}, it is not straightforward in our matching formulation (Eq.~(\ref{eqn:matching})) to associate both features at different levels.
To enable our model to capture an arbitrary correspondence between image and label features, we design a hierarchical architecture that adaptively relates different levels of image and label features depending on each task.

\begin{figure}[t!]
    \centering
    \includegraphics[width=0.8\linewidth]{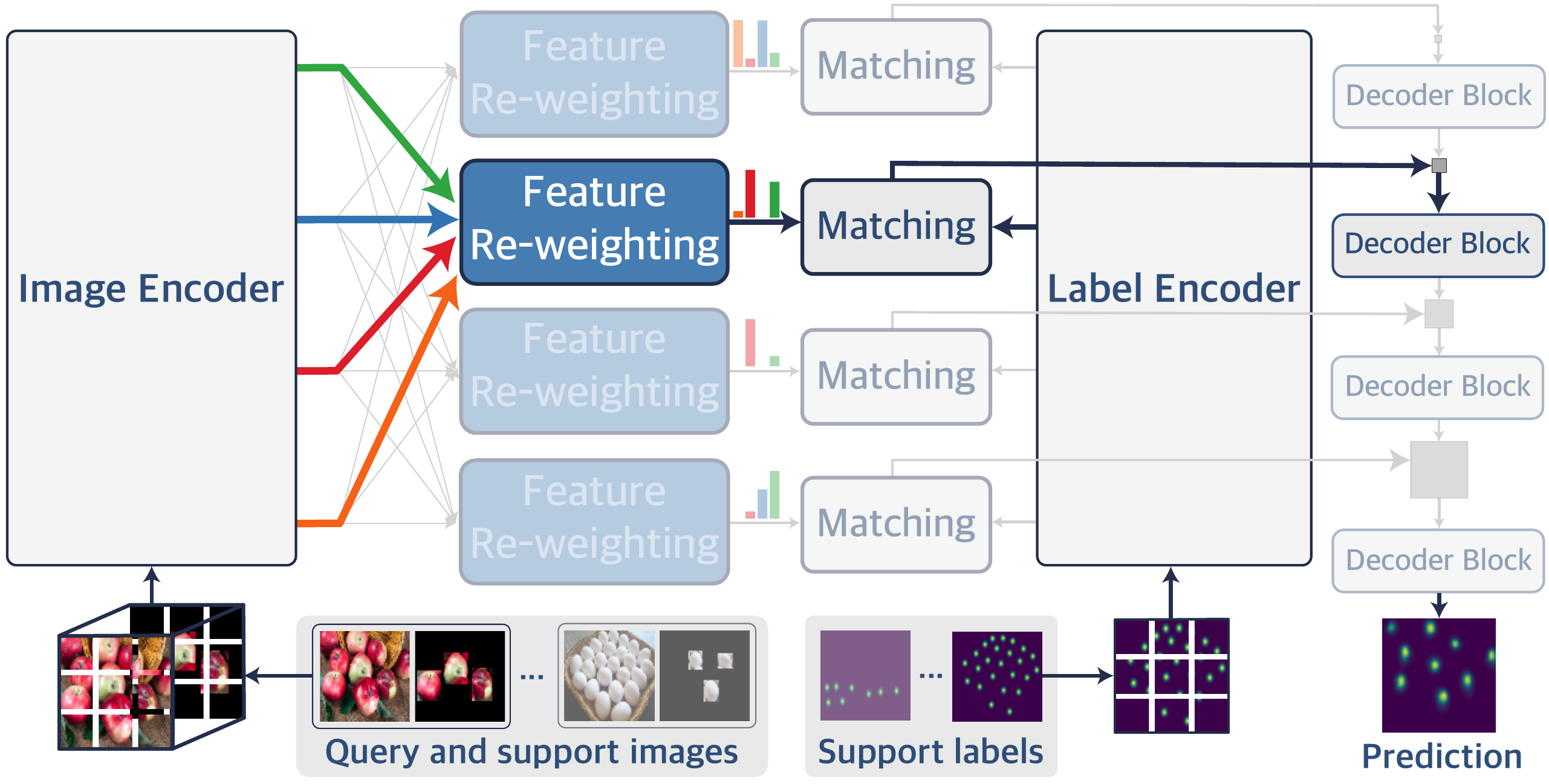}
    \caption{
    Task-adaptive feature re-weighting mechanism with a hierarchical architecture.
    The figure highlights the matching module at the third level of the hierarchy ($l=3$).
    }
    \label{fig:feature_reweighting}
    \vspace{-0.4cm}
\end{figure}

To this end, we extract the image and label features at $L$ levels of their encoders and perform matching at each label feature level using all levels of the image features, as illustrated in Figure~\ref{fig:feature_reweighting}.
To control the contribution of the image feature levels on each matching module task-specifically, we introduce a learnable matrix $\Lambda_\mathcal{T} \in \mathbb{R}^{L \times L}$ for each task $\mathcal{T}$ that re-weights the multi-level image features $\hat{F}_\mathcal{T} = [\hat{f}^{(1)}_\mathcal{T}(\mathbf{x}), \cdots, \hat{f}^{(L)}_\mathcal{T}(\mathbf{x})] \in \mathbb{R}^{L \times d}$ via matrix multiplication:
\begin{equation}
    F_\mathcal{T} = [f^{(1)}_\mathcal{T}(\mathbf{x}), \cdots, f^{(L)}_\mathcal{T}(\mathbf{x})] = \Lambda_\mathcal{T} \hat{F}_\mathcal{T},
\end{equation}
where each row of $\Lambda_\mathcal{T}$ is normalized to sum to 1 such that the total contribution from the image feature levels remains constant.
Then each re-weighted feature $f^{(l)}_\mathcal{T}(\mathbf{x})$ is passed to $l$-th matching module:
\begin{equation}
    g^{(l)}\left(\mathbf{y}_k^q\right) = \sum_{i \le N} \sum_{j \le M} \sigma^{(l)} \left( f^{(l)}_\mathcal{T}\left(\mathbf{x}_k^q\right), f^{(l)}_\mathcal{T}\left(\mathbf{x}_j^i\right) \right) \cdot g^{(l)}\left(\mathbf{y}_j^i\right), \quad 1 \le l \le L.
\label{eqn:matching_reweighted}
\end{equation}
After performing the matching at $L$ levels, we convert the outputs into a feature pyramid whose resolution increases as the level decreases, which are progressively decoded by a convolutional decoder~\cite{ranftl2021vision}.

In this way, our model can adapt to various tasks having different optimal correspondence between the image and label features (see Figure~\ref{fig:feature_weights} for the learned feature weights in downstream tasks), as well as adapting the image features themselves via bias tuning.
Since the task-specific parameters introduced in the image encoder $\theta_\mathcal{T} = (P_\mathcal{T}, \mathbf{b}_\mathcal{T}, \Lambda_\mathcal{T})$ occupy a small portion of the whole parameters, \modelname{} is robust to over-fitting during fine-tuning.

\section{Scaling up the Data and the Model}
\label{sec:training}
\vspace{-0.3cm}

We investigate strategies to enhance the generalization of \modelname{} over various unseen dense prediction tasks by collecting a large-scale meta-training dataset (Section~\ref{sec:scaleup_data}) and scaling up the model capacity and resolutions (Section~\ref{sec:scaleup_model}).

\cutsectionup
\subsection{Meta-Training Data with Diverse Tasks and Domains}
\label{sec:scaleup_data}
\vspace{-0.2cm}

\begin{figure}[t!]
    \centering
    \begin{subfigure}{.45\textwidth}
      \centering
        \includegraphics[width=\linewidth]{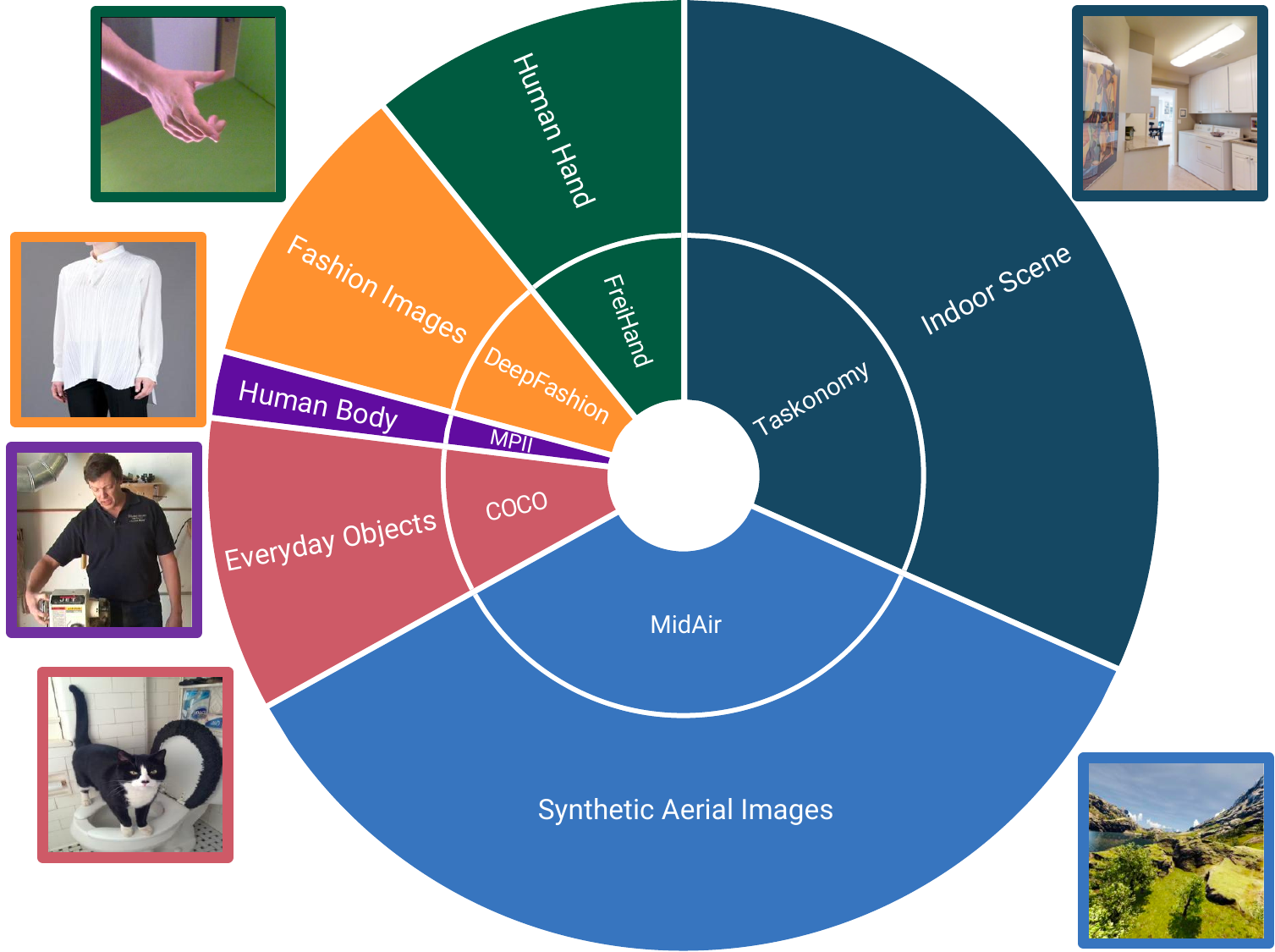}
    \end{subfigure}%
    \hspace{0.06\textwidth}
    \begin{subfigure}{.45\textwidth}
      \centering
      \includegraphics[width=\linewidth]{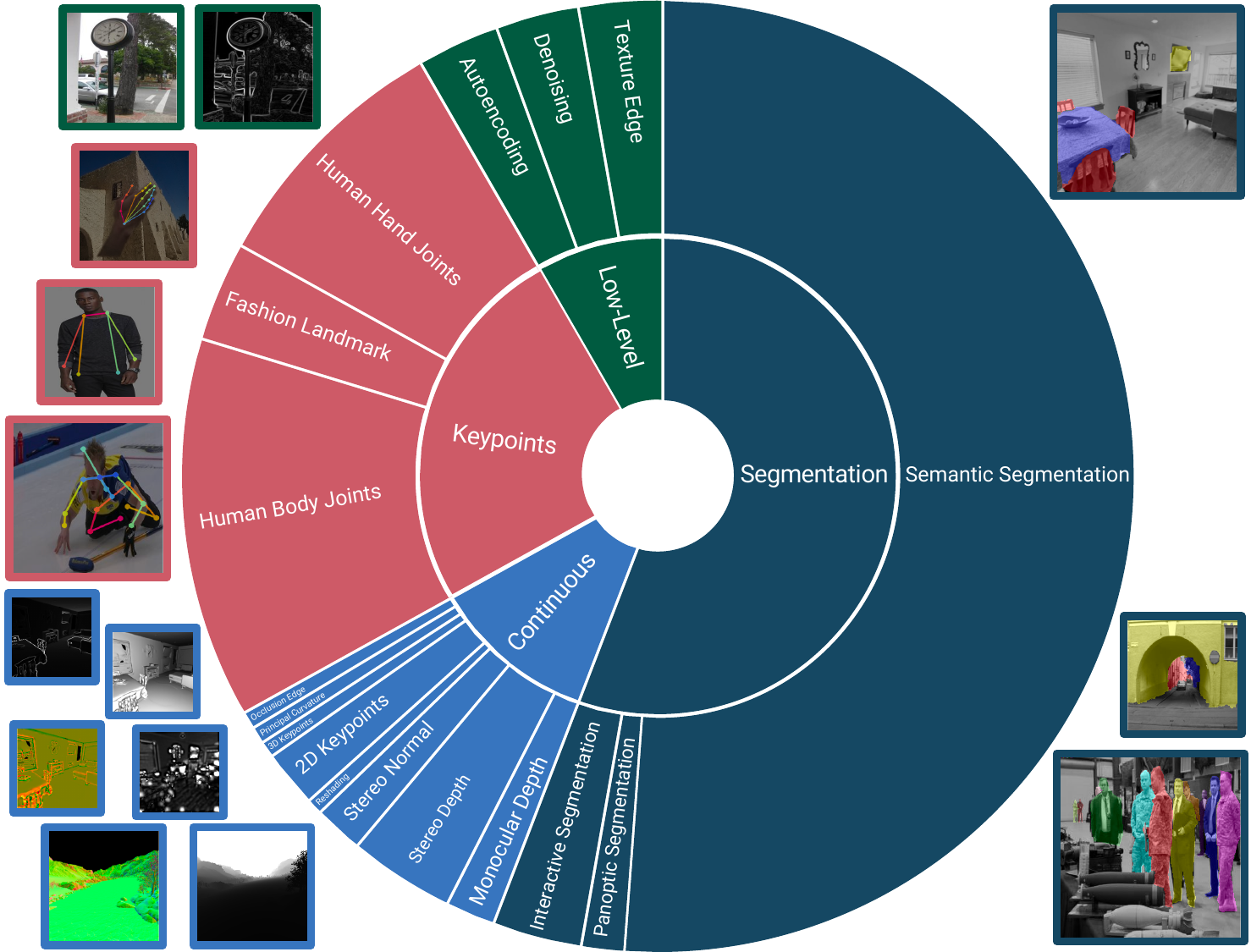}
    \end{subfigure}
    \caption{Summary of our meta-training dataset. Left: image domains (outer circle) and source datasets (inner circle). Sizes correspond to the dataset size. Right: task categories (inner circle) and specific tasks (outer circle). Sizes correspond to the sampling ratio.}
    \label{fig:data_chart}
    \vspace{-0.5cm}
\end{figure}

For achieving robust generalization in dense visual prediction tasks across real-world scenarios, meta-training on diverse domains and tasks constitutes a crucial element of \modelname{}. 
To this end, we curated a large-scale meta-training dataset comprising around 1.2 million images drawn from six prominent datasets: Taskonomy~\cite{zamir2018taskonomy}, COCO~\cite{lin2014microsoft,caesar2018coco}, MidAir~\cite{Fonder2019MidAir}, MPII~\cite{andriluka14cvpr}, DeepFashion~\cite{liuLQWTcvpr16DeepFashion}, and FreiHand~\cite{Freihand2019}.
As summarized in Figure~\ref{fig:data_chart}, the dataset covers a wide range of domains (indoor to outdoor) and tasks (continuous to categorical) considered in mainstream vision benchmarks, which makes \modelname{} generally applicable to many real-world scenarios.

Our meta-training dataset consists of dense labels from 14 different dense prediction tasks, which can be roughly categorized into continuous signal prediction, semantic segmentation, and keypoint detection.
We also augment the dataset with three unsupervised tasks, namely autoencoding, denoising, and edge detection (see Figure~\ref{fig:data_chart} for the sampling ratio of each task).
To include tasks with multi-modal input, we use stereo images offered by the MidAir dataset.
In addition, we simulate an interactive segmentation task using instance segmentation labels in the COCO dataset by composing a pair of images as input, where the first element is an RGB image and the second image includes marked positions of several pixels sampled within the target object instances to be segmented.

\cutparagraphup
\subsection{Scaling up the Model}
\vspace{-0.1cm}
\label{sec:scaleup_model}
To boost the performance of \modelname{} in the wild, we scale up the model capacity from a base implementation of VTM~\cite{kim2023universal}.
Since the image encoder plays a central role in the matching, we scale it up to pre-trained BEiTv2-Large~\cite{peng2022beit}.
To match the correspondence between the image and label encoders, we also scale up the label encoder to ViT-Large~\cite{dosovitskiy2021an} and increase the dimension and number of heads in the matching module accordingly. 
Finally, the number of convolution channels in the label decoder has increased from 96 to 256.

Since the performance of dense prediction is generally sensitive to the image resolution, \modelname{} adapts to the resolution $(H_\mathcal{T}, W_\mathcal{T})$ defined for each target task $\mathcal{T}$.
This can be done by performing spatial interpolation of the positional embeddings of the transformer encoders, both for images and labels.
To avoid the heavy meta-training at high resolution, we meta-train \modelname{} with $(224, 224)$ resolution and then fine-tune it with the adapted resolution, which efficiently boosts up the downstream performance.
\cutparagraphup
\section{Experiments}
\label{sec:experiments}

\vspace{-0.2cm}
This section presents the evaluation results of \modelname{} on six benchmark datasets and internal analysis. 
More results and detailed descriptions of implementation and experiments are in the Appendix.

\paragraph{Generalist Baselines}
We compare our model with three data-efficient generalist approaches: Painter~\cite{wang2022images}, SegGPT~\cite{wang2023seggpt}, and VTM~\cite{kim2023universal}.
Painter and SegGPT can be used in unseen tasks with or without test-time adaptation through In-Context Learning (ICL) or Prompt Tuning (PT), respectively.
Therefore, we evaluate Painter and SegGPT under both settings, where we apply SegGPT+ICL in only segmentation tasks since the model cannot handle the continuous label. 
For a fair comparison, we use the same support set for fine-tuning (VTM, Painter+PT, SegGPT+PT) and prompting (Painter+ICL, SegGPT+ICL).
As all of these baselines do not support multiple input images, we apply them on tasks having a single input image.

\paragraph{Specialist Baselines}
To provide a reference, we also report the performance of two specialist models for each task trained with full supervision.
Since our goal is \emph{not} beating the state-of-the-arts in individual benchmarks but demonstrating the generality, we avoid specialists that incorporate heavy task-specific post-processing or extra supervision as they are orthogonal to the model.

\cutparagraphup
\subsection{Downstream Tasks}
\vspace{-0.2cm}
To evaluate the generality of our method in real-world few-shot settings, we select six downstream tasks covering diverse output semantics and structures as well as input domains and modalities that are \textbf{unseen} in the meta-training.

\paragraph{Animal Keypoint Detection}
To test whether our model can flexibly adapt to unseen output structures, we select animal keypoint detection.
The objective is to predict the joint locations of animals, which can be converted to a multi-channel dense heatmap.
Note that the output structure, \emph{i.e.,} definition of keypoints and their spatial relationships, are unseen during meta-training.
We evaluate our model on the AP-10K~\cite{yu2021ap} dataset, where we select eight species with distinctive features (antelope, cat, elephant, giraffe, hippo, horse, mouse, and pig) and report the mean average precision (AP)~\cite{yu2021ap} over them.
For simplicity of post-processing, we exclude images with multiple instances.

\paragraph{6D Pose Estimation}
To test whether our model can also adapt to unseen output semantics, we select 6D pose estimation.
The objective is to predict the 6D extrinsic camera matrix that represents the rotation and translation of a target object.
We formulate it as a dense prediction by predicting dense correspondence between each image pixel and 3D vertex of the provided CAD model, from which the 6D pose is obtained by Perspective-n-Point algorithm~\cite{lepetit2009ep}.
Indeed, the labels have distinct semantics and structure from those of meta-training tasks.
We evaluate our model on the LineMOD~\cite{hinterstoisser2011linemod} dataset and report the ADD score~\cite{rad2017bb8} measuring the distance of vertices in 3D space.

\paragraph{Exemplar-Guided Object Counting}
To test whether our model can exploit a user interaction as an extra image modality, we select exemplar-guided object counting.
The objective is to count all objects in an image specified by three bounding box exemplars, which are represented by two images: RGB image and an exemplar guide that highlights the bounding box areas.
In this task, the model must use the exemplar guide to figure out target objects to be counted.
We formulate the task to predict the heatmap of object centers, from which the number of objects is obtained by counting the modes.
We employ the FSC-147~\cite{ranjan2021fsc} dataset and report mean absolute error (MAE) following the literature~\cite{liu2022countr,djukic2023low}.

\paragraph{Cell Instance Segmentation}
Cell instance segmentation also has multi-modal input images, with distinct domains from the natural images.
The objective of this task is to segment all cell instances within a bi-model image (one for cytoplasm and another for nuclei).
Following \cite{stringer2021cellpose}, we formulate the task as flow estimation, where the model predicts vertical and horizontal gradients of each cell towards its center along with foreground segmentation.
As in 6D pose estimation, this output representation has distinct semantics and structures from the meta-training tasks.
We evaluate our model on the Cellpose~\cite{stringer2021cellpose} dataset and report average precision with threshold IoU=0.5 ($\text{AP}_{50}$).

\paragraph{Skin Lesion Segmentation}
We select skin lesion segmentation as an \emph{in-distribution} but \emph{out-of-domain} task, where the objective is to segment the skin lesion in dermatoscopic images.
We employ ISIC 2018~\cite{milton2019isic} dataset and report average F1 score of 5-fold cross-validation, following the literature~\cite{he2022fully,valanarasu2022unext}.

\paragraph{Video Object Segmentation}
Finally, to further explore the potential of our model in the wild, we select video object segmentation.
The objective is to track target objects over an entire video, which are specified in the first frame.
We formulate this task as 1-shot image segmentation by treating the first frame as support and the remaining as queries, where we augment the 1-shot support with random cropping.
Note that, unlike common specialists in this literature, we neither exploit any temporal correlation nor train our model on video data.
We employ the DAVIS 2017~\cite{jordi2017davis} dataset and report the $\mathcal{J}$\&$\mathcal{F}$ score~\cite{cheng2022xmem,wang2023look}.

\begin{figure*}[!t]
    \begin{minipage}{\textwidth}
    \captionsetup{type=table}
    \centering
    \caption{Comparison with specialists for each task and generalists based on in-context learning and parameter-efficient fine-tuning. Generalists use 1-shot support for DAVIS 2017, 20-shot for AP-10K and ISIC 2018, and 50-shot for the others.}
    \vspace{-0.3cm}
    \label{tab:main_table}
    \begin{center}
        \renewcommand{\arraystretch}{1.2}
        \renewcommand{\aboverulesep}{0pt}
        \renewcommand{\belowrulesep}{0pt}
        \setlength\tabcolsep{1pt}
        \scriptsize
        \begin{tabular}{c | c p{0.01cm} c p{0.01cm} c p{0.01cm} c p{0.01cm} c p{0.01cm} c}
            \toprule
            \multirow{3}{*}{\qquad} &
            
            animal kp. & &
            6D pose & &
            skin les. seg. & &
            video obj. seg. & &
            obj. count. & &
            cell inst. seg.
            \\
    
            &
            AP-10K & &
            LineMOD & &
            ISIC 2018 & &
            DAVIS 2017 & &
            Cellpose & &
            FSC-147
            \\
    
            \cmidrule{2-12}
            
            &
            AP ↑ & &
            ADD ↑ & &
            F1 ↑ & &
            $\mathcal{J}$\&$\mathcal{F}$ ↑ & &
            MAE ↓ & &
            $\text{AP}_{50}$ ↑
            \\
    
            \midrule
    
            \multicolumn{12}{c}{Specialists, fully-supervised}
            \\
            
            \midrule
    
            SimpleBaseline~\cite{xiao2018simple} &
            64.9 & &
            - & &
            - & &
            - & &
            - & &
            -
            \\
            
            HRNet~\cite{sun2019deep} &
            69.8 & &
            - & &
            - & &
            - & &
            - & &
            -
            \\
    
            DPOD~\cite{zakharov2019dpod} &
            - & &
            83.0 & &
            - & &
            - & &
            - & &
            -
            \\
    
            CDPN~\cite{li2019cdpn} &
            - & &
            89.9 & &
            - & &
            - & &
            - & &
            -
            \\
    
            FTN~\cite{he2022fully} &
            - & &
            - & &
            89.7 & &
            - & &
            - & &
            -
            \\
    
            UNeXt~\cite{valanarasu2022unext} &
            - & &
            - & &
            89.8 & &
            - & &
            - & &
            -
            \\
    
            XMem~\cite{cheng2022xmem} &
            - & &
            - & &
            - & &
            87.7 & &
            - & &
            -
            \\
    
            ISVOS~\cite{wang2023look} &
            - & &
            - & &
            - & &
            88.2 & &
            - & &
            -
            \\
    
            CounTR~\cite{liu2022countr} &
            - & &
            - & &
            - & &
            - & &
            12.0 & &
            -
            \\
    
            LOCA~\cite{djukic2023low} &
            - & &
            - & &
            - & &
            - & &
            10.8 & &
            -
            \\
    
            Stardist~\cite{schmidt2018cell} &
            - & &
            - & &
            - & &
            - & &
            - & &
            67.0
            \\
    
            Cellpose~\cite{stringer2021cellpose} &
            - & &
            - & &
            - & &
            - & &
            - & &
            70.4
            \\
    
            \midrule
    
            \multicolumn{12}{c}{Generalists, low-shot}
            \\
            
            \midrule
            
            Painter~\cite{wang2022images} + ICL &
            0 & &
            0 & &
            36.3 & &
            34.6 & &
            - & &
            -
            \\
            
            Painter~\cite{wang2022images} + PT &
            0.4 & &
            15.7 & &
            86.8 & &
            53.9 & &
            - & &
            -
            \\
    
            SegGPT~\cite{wang2023seggpt} + ICL &
            - & &
            - & &
            60.2 & &
            75.6 & &
            - & &
            -
            \\
    
            SegGPT~\cite{wang2023seggpt} + PT &
            2.0 & &
            23.1 & &
            88.1 & &
            67.0 & &
            - & &
            -
            \\
            
            VTM~\cite{kim2023universal} &
            9.1 & &
            59.3 & &
            84.1 & &
            54.0 & &
            - & &
            -
            \\
    
            \midrule
            
            \textbf{\modelname{} (ours)} &
            67.2 & &
            85.2 & &
            88.5 & &
            77.5 & &
            12.0 & &
            70.3
            \\
    
            \bottomrule
        \end{tabular}
        \vspace{0.1cm}
    \end{center}
    \end{minipage}
    \begin{minipage}{\textwidth}
    \centering
    \includegraphics[width=\linewidth]{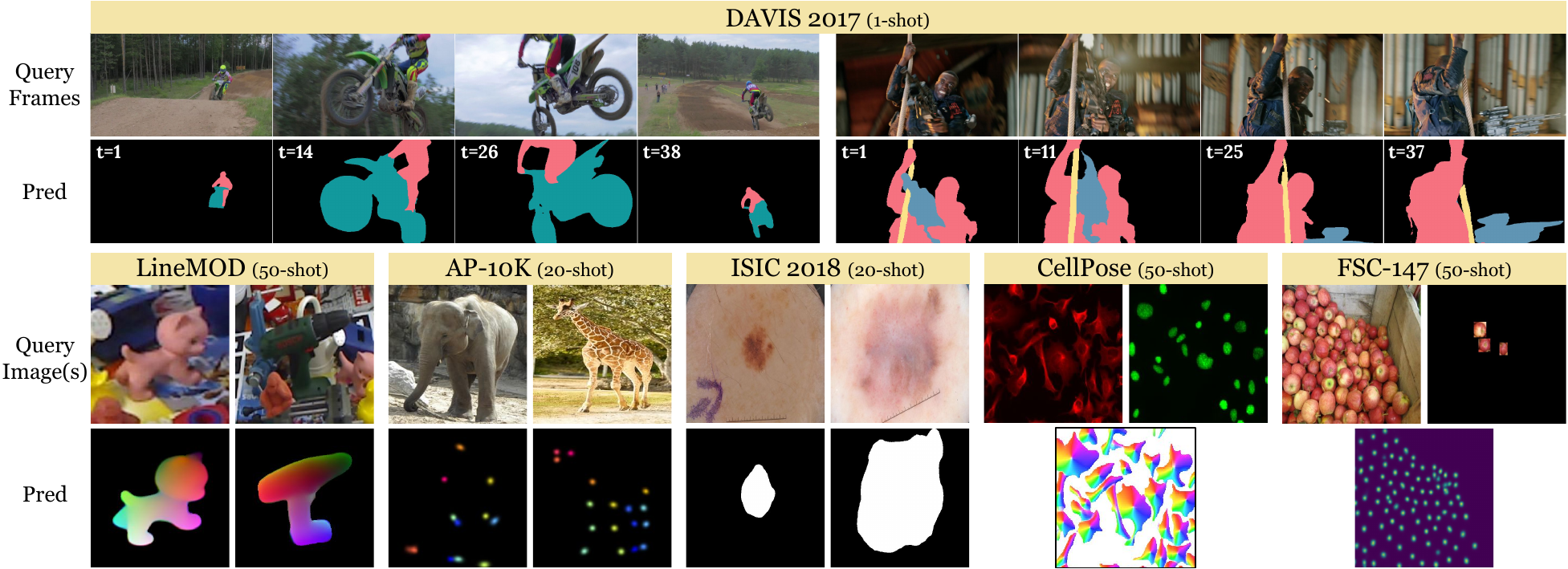}
    \captionsetup{type=figure}
    \vspace{-0.5cm}
    \caption{Qualitative results of \modelname{} in six downstream benchmarks.
    We color-coded outputs from different channels. $t$ denotes the frame number.
    }
    \label{fig:main_qualitative}
    \vspace{-0.5cm}

    \end{minipage}
\end{figure*}

\vspace{-0.5cm}
\subsection{Main Results}
\vspace{-0.2cm}
Table~\ref{tab:main_table} summarizes the performance of our model and baselines on the six downstream tasks.
In general, our model significantly outperforms the generalist baselines in all tasks, which shows the effectiveness of our approach in low-shot learning of diverse dense visual prediction in real-world applications.
We discuss the results of each task in the following paragraphs.

\begin{figure}[t!]
    \centering
    \includegraphics[width=\linewidth]{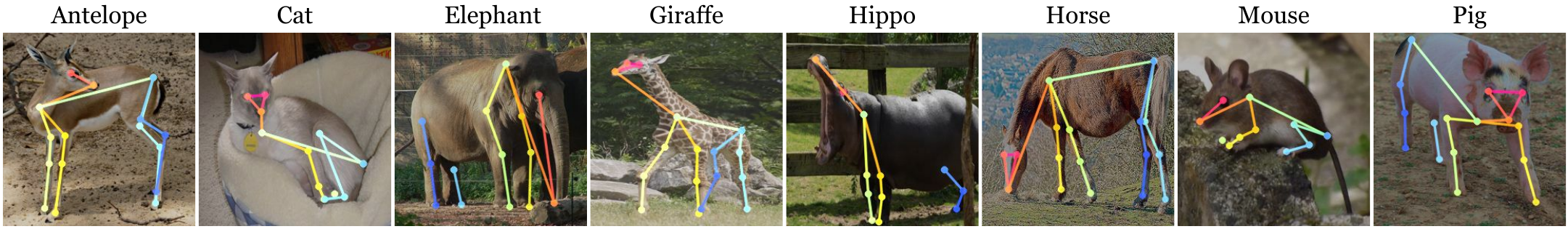}
    \vspace{-0.5cm}
    \caption{Keypoint prediction of \modelname{} on eight animal species.}
    \label{fig:ap10k_qualitative}
    \vspace{-0.2cm}
\end{figure}

\paragraph{Animal Keypoint Detection}
In this task, our model should understand not only the appearance of distinctive animal body parts but also their spatial priors to resolve ambiguities in prediction.
Since these are largely different across species and any objects in meta-training data, the task requires rapid adaptation to unseen domains and output structures.
As shown in Figure~\ref{fig:ap10k_qualitative}, \modelname{} successfully predicts keypoints of eight species with varying appearance and body configurations.
Interestingly, our model seems to leverage the spatial prior to localizing the missing parts (occlusions in Antelope and Cat) and distinguish left and right, showing effectiveness in adaptation.
We also observe that the generalist baselines struggle to learn this task despite the test-time adaptation (see Figure~\ref{fig:oodgen_comparison}), showing the effectiveness of our model in adapting to unseen output structures.

\begin{figure}[t!]
    \centering
    \includegraphics[width=\linewidth]{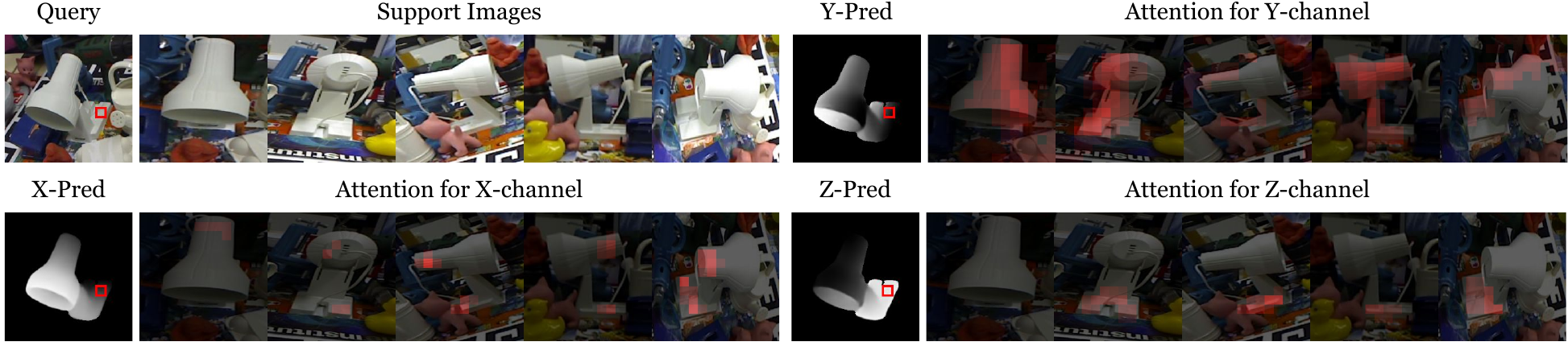}
    \vspace{-0.45cm}
    \caption{
    Visualization of attention maps between a query (red box in the first image) and support patches in output channels of 6D pose estimation. 
    \modelname{} captures 3D relationship between query and support by attending back, left, and bottom parts of the object for X, Y, and Z channels, respectively.
    }
    \label{fig:linemod_attention}
    \vspace{-0.5cm}
\end{figure}

\paragraph{6D Pose Estimation}
In this task, our model has to predict 6D pose of an object, which is different from any meta-training tasks in both knowledge to solve it and the output structure.
Without leveraging a dedicated architecture for 3D understanding, \modelname{} successfully adapts to the task, even outperforming some of the specialized baselines.
To further analyze if our model really understands the task, we visualize the attention score in the matching (Eq.~(\ref{eqn:matching})) in Figure~\ref{fig:linemod_attention}.
It shows that the similarity of the query image patch with the support images is highly correlated with 3D positions, which is desirable for the task.
We also note that learned weights in the feature re-weighting (Figure~\ref{fig:feature_weights}) tend to be inversely correlated with the feature levels. 
It indicates that the model leverages the high-level semantics to capture fine details in labels, which is reasonable in 3D understanding.
These observations indicate that our model can adapt to novel 3D understanding tasks with unique output structure.

\begin{figure}[t!]
    \centering
    \includegraphics[width=1.0\linewidth]{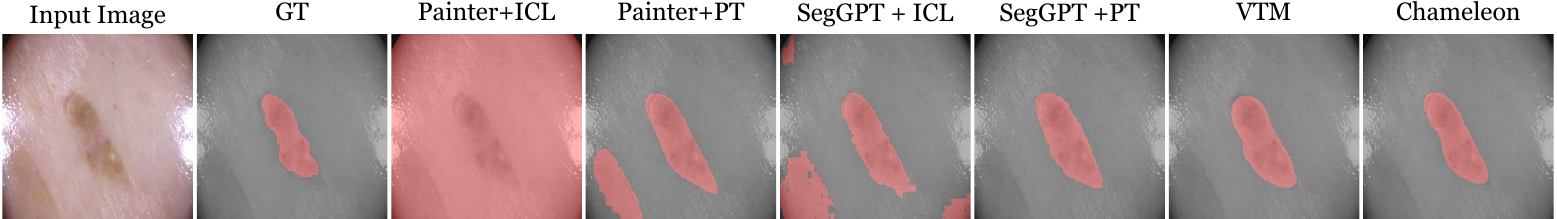}
    \vspace{-0.4cm}
    \caption{Qualitative comparison between generalist models in out-of-domain prediction of medical semantic segmentation.}
    \label{fig:isic_qualitative}
    \vspace{-0.3cm}
\end{figure}

\paragraph{Medical Semantic Segmentation}
In this task, the model has to adapt to a huge domain shift from natural images in meta-training data to medical images.
As shown in Figure~\ref{fig:isic_qualitative} and Table~\ref{tab:main_table}, our model successfully adapts even with such domain gaps, while in-context learning methods struggle.
Not surprisingly, with prompt tuning, Painter and SegGPT become competitive with our model, as they can address out-of-domain tasks with seen label semantics and structures.
Still, \modelname{} outperforms all the generalist baselines, showing its effectiveness.

\begin{figure}[t!]
    \centering
    \includegraphics[width=0.85\linewidth]{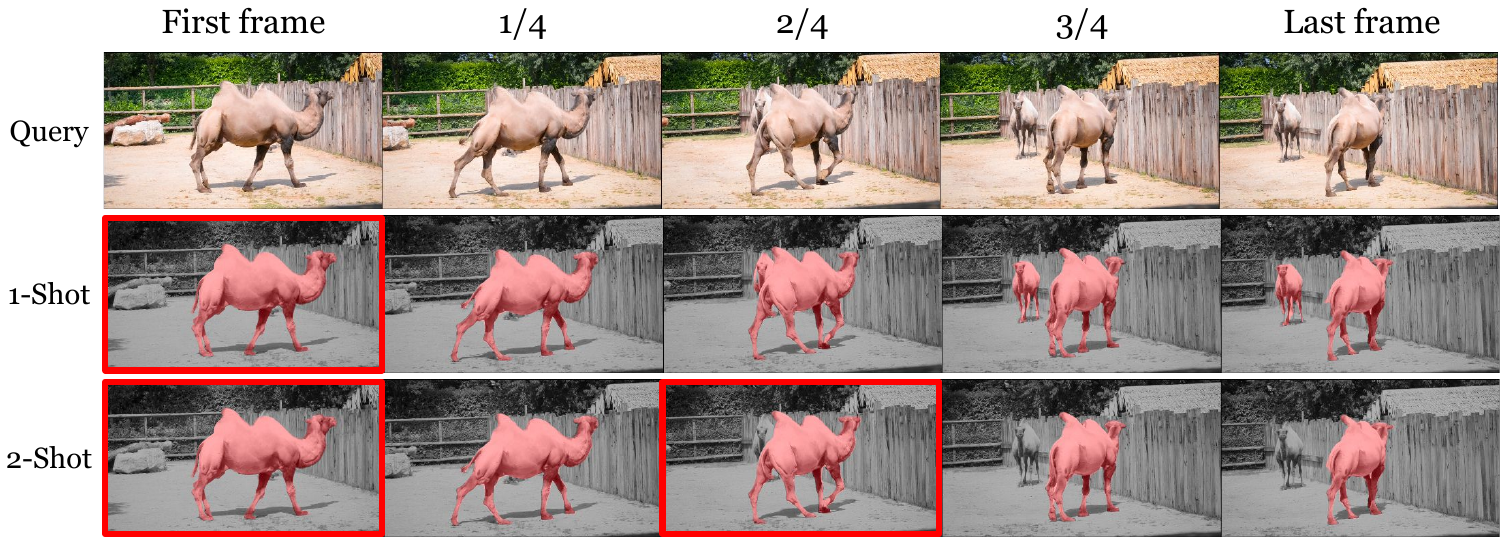}
    \vspace{-0.2cm}
    \caption{
    Video segmentation results with one and two support frames (red boxes). 
    }
    \label{fig:davis_2shot}
    \vspace{-0.5cm}
\end{figure}

\paragraph{Video Object Segmentation}
Although the segmentation is a part of the meta-training tasks, generalizing our model to video object segmentation is challenging since it is learned on images and unaware of relating temporally distant objects.
Surprisingly, by matching each frame independently with the label of the first frame, \modelname{} successfully tracks objects under significant appearance variations (Figure~\ref{fig:main_qualitative}), achieving comparable performance to the supervised methods that heavily rely on temporal correlation.
As shown in Figure~\ref{fig:davis_2shot}, most failure cases of our method are due to ambiguous distractors, which can be resolved by additional frame labels.
Indeed, our method can naturally incorporate such additional labels while it is not straightforward in baselines due to the causal inference, and \modelname{} begins to surpass the specialists with four frame labels (Figure~\ref{fig:performance_on_shots}).

\begin{figure}[t!]
    \centering
    \includegraphics[width=.89\linewidth]{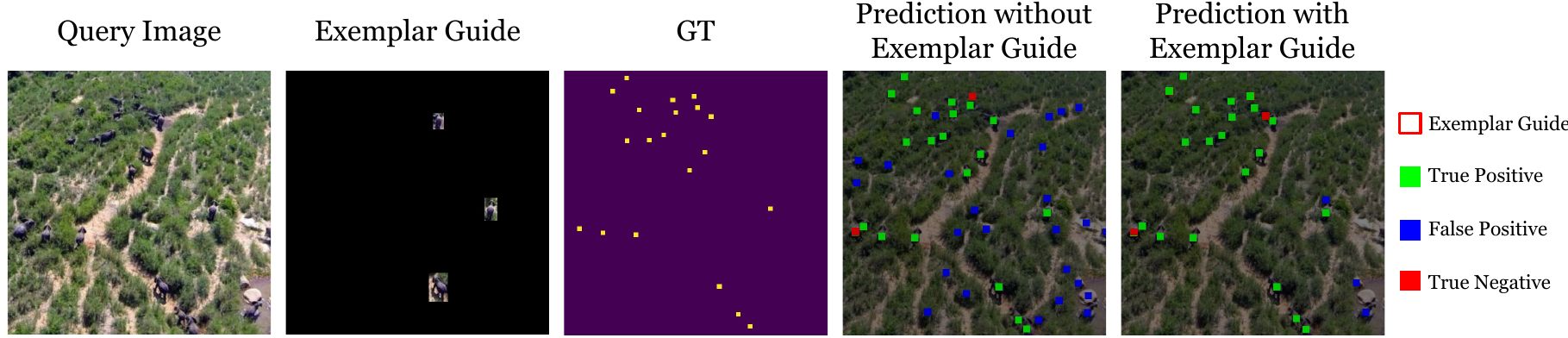}\\
    {\footnotesize
    (a) Effect of using an exemplar guide in object counting.}
    \\
    \vspace{0.15cm}
    \includegraphics[width=.89\linewidth]{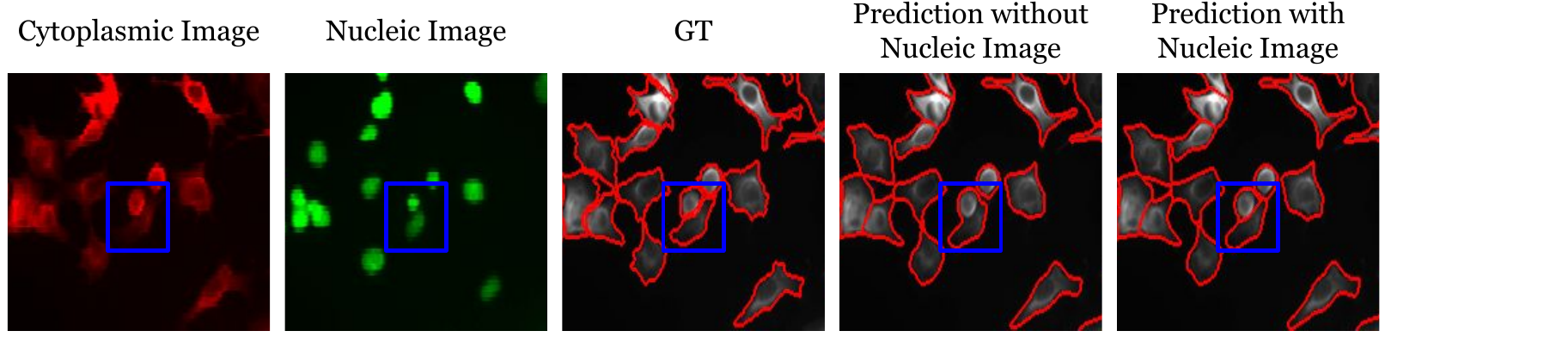}\\
    {\footnotesize
    (b) Effect of using a nucleic image in cell instance segmentation.}
    \vspace{-0.2cm}
    \caption{Effect of using multi-modal input. (a) In object counting, \modelname{} excludes false positives (bushes) by using the exemplar guide. (b) In cell instance segmentation, \modelname{} separates two cells in the blue box by exploiting the nucleic image.}
    \label{fig:multimodal_qualitative}
    \vspace{-0.5cm}
\end{figure}

\paragraph{Exemplar-Guided Object Counting}
The ability to process multiple input images also allows \modelname{} to be applied to user-interactive tasks, such as exemplar-guided object counting.
In this task, our model exploits the exemplar guide given as the second image to identify objects to count.
As shown in Figure~\ref{fig:multimodal_qualitative}~(a), counting objects without such guidance inevitably includes many false positives, whereas our method successfully excludes them using the guidance.
Together with cell instance segmentation tasks, it shows that our method can adapt to multi-modal inputs with vastly different semantics effectively with the encoding mechanism introduced in Section~\ref{subsec:multi_input}.

\paragraph{Cell Instance Segmentation}
This task involves out-of-domain images and labels, but more interestingly, solving this task requires understanding of bi-modal images of cytoplasm and nuclei.
It requires our model to take these two images and learn to leverage their exclusive cues for cell instance segmentation by adapting the multi-modal position bias.
As shown in Figure~\ref{fig:multimodal_qualitative}~(b), \modelname{} successfully utilizes such information, by distinguishing two instances entangled in cytoplasmic image using information in nucleic image.

\vspace{-0.3cm}
\subsection{Ablation Study}
\vspace{-0.2cm}

\begin{figure}[t!]
    \centering
    \includegraphics[width=\linewidth]{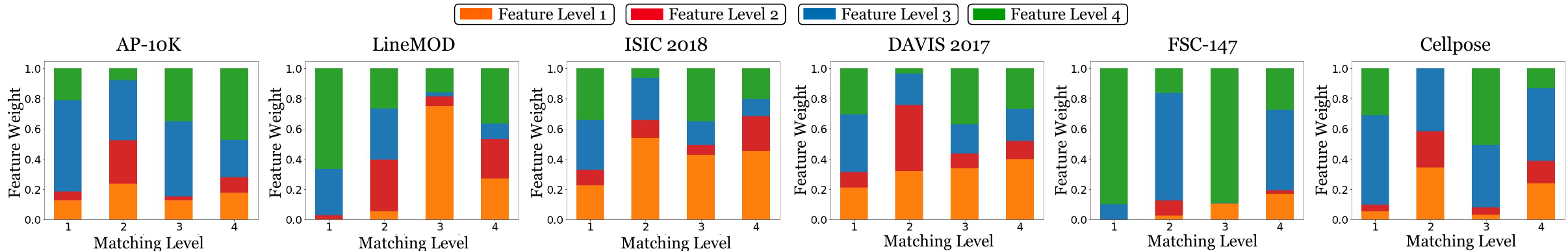}
    \vspace{-0.6cm}
    \caption{Learned feature weights in downstream tasks.}
    \label{fig:feature_weights}
\end{figure}

\paragraph{Component-wise Analysis}
We conduct an ablation study to analyze the effect of each component introduced in \modelname{}.
As our model is based on the VTM framework, we ablate our improvements from VTM one by one.
As shown in Table~\ref{tab:model_ablation_table}, all components contribute to improving the downstream performance (scaling up the model and meta-training data), as well as broadening the scope to multi-modal applications (encoding mechanism for multi-modal inputs).
Notably, we observe that feature re-weighting improves the performance considerably especially when the structure and semantics of the labels are largely different from meta-training tasks, such as 6D pose estimation, object counting, and cell instance segmentation.
As shown in Figure~\ref{fig:feature_weights}, the learned weights vary substantially across tasks, showing its effectiveness in adapting matching modules to the out-of-distribution tasks.

\paragraph{Ablation Study on Meta-Training Data}
To further analyze the effect of meta-training, we conduct an ablation study in Table~\ref{tab:data_ablation_table} by gradually increasing the scale and diversity of meta-training data.
The performance of \modelname{} tends to consistently improve as we diversify domains and tasks in the meta-training dataset. 
Interestingly, such improvements are often from adding tasks \emph{less or barely correlated} with the downstream tasks.
For instance, adding synthetic drone images with continuous labels (MidAir) improves the animal pose estimation by a large margin, or adding keypoint detection tasks (KP-4) improves 6D pose estimation.
It shows that our method can effectively leverage the indirect correlations of meta-training and downstream tasks through universal matching, which is critical in generalization to unseen out-of-distribution tasks.

\begin{table*}[!t]
\caption{Ablation study on the contributions of each component.}
\vspace{-0.8cm}
\label{tab:model_ablation_table}
\begin{center}
    \renewcommand{\arraystretch}{1.2}
    \renewcommand{\aboverulesep}{0pt}
    \renewcommand{\belowrulesep}{0pt}
    \setlength\tabcolsep{2pt}
    \scriptsize
    \begin{tabular}{c | c p{0.01cm} c p{0.01cm} c p{0.01cm} c p{0.001cm} c p{0.001cm} c}
        \toprule
        \multirow{2}{*}{Model Variant} &
        
        AP-10K & &
        LineMOD & &
        ISIC 2018 & &
        DAVIS 2017 & &
        FSC-147 & &
        Cellpose
        \\

        \cmidrule{2-12}
        
        &
        AP ↑ & &
        ADD ↑ & &
        F1 ↑ & &
        $\mathcal{J}$\&$\mathcal{F}$ ↑ & &
        MAE ↓ & &
        $\text{AP}_{50}$ ↑
        \\

        \midrule
        
        VTM &
        9.1 & &
        59.3 & &
        84.1 & &
        54.0 & &
        - & &
        -
        \\
        
        + Enlarged Backbone &
        25.3 & &
        73.9 & &
        85.4 & &
        66.8 & &
        - & &
        -
        \\

        + Large and Diverse Data &
        65.0 & &
        73.6 & &
        86.5 & &
        73.6 & &
        - & &
        -
        \\

        + Variable-Input Encoder &
        63.6 & &
        73.3 & &
        \textbf{88.9} & &
        76.0 & &
        17.9 & &
        67.2
        \\

        + Feature Weighting \textbf{(Ours)} &
        \textbf{67.2} & &
        \textbf{85.2} & &
        88.5 & &
        \textbf{77.5} & &
        \textbf{12.3} & &
        \textbf{70.3}
        \\

        \bottomrule
    \end{tabular}
\end{center}
\vspace{-0.1cm}

\caption{Ablation study on meta-training dataset. COCO (seg.) refers to using only segmentation labels in COCO dataset, and KP-4 refers to using four keypoint detection datasets (COCO, MPII, Deepfashion, and Freihand).}
\vspace{-0.7cm}
\label{tab:data_ablation_table}
\begin{center}
    \renewcommand{\arraystretch}{1.2}
    \renewcommand{\aboverulesep}{0pt}
    \renewcommand{\belowrulesep}{0pt}
    \setlength\tabcolsep{1.3pt}
    \scriptsize
    \begin{tabular}{cccc | c p{0.01cm} c p{0.01cm} c p{0.01cm} c p{0.01cm} c p{0.01cm} c}
        \toprule
        \multirow{2}{*}{Taskonomy} &
        \multirow{2}{*}{MidAir} &
        \multirow{2}{*}{\shortstack{COCO\\(seg.)}} &
        \multirow{2}{*}{KP-4} &

        AP-10K & &
        LineMOD & &
        ISIC 2018 & &
        DAVIS 2017 & &
        FSC-147 & &
        Cellpose
        \\

        \cmidrule{5-15}
        
        & & & &
        AP ↑ & &
        ADD ↑ & &
        F1 ↑ & &
        $\mathcal{J}$\&$\mathcal{F}$ ↑ & &
        MAE ↓ & &
        $\text{AP}_{50}$ ↑
        \\

        \midrule
        
        \cmark & \xmark & \xmark & \xmark &
        19.3 & &
        84.0 & &
        84.6 & &
        66.4 & &
        - & &
        -
        \\
        
        \cmark & \cmark & \xmark & \xmark &
        42.4 & &
        80.0 & &
        87.3 & &
        69.3 & &
        17.1 & &
        66.9
        \\

        \cmark & \cmark & \cmark & \xmark &
        65.1 & &
        83.3 & &
        88.0 & &
        74.6 & &
        14.9 & &
        69.2
        \\

        \cmark & \cmark & \cmark & \cmark &
        \textbf{67.2} & &
        \textbf{85.2} & &
        \textbf{88.5} & &
        \textbf{77.5} & &
        \textbf{12.3} & &
        \textbf{70.3}
        \\

        \bottomrule
    \end{tabular}
\vspace{-0.4cm}
\end{center}
\end{table*}

\begin{figure}[t!]
    \centering
    \includegraphics[width=\linewidth]{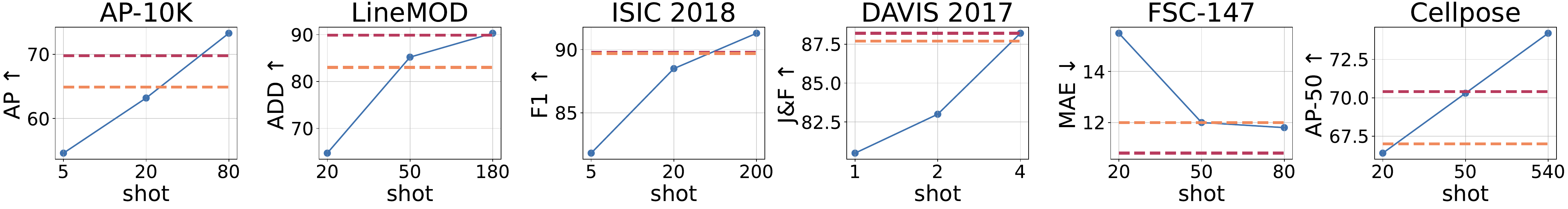}
    \vspace{-0.6cm}
    \caption{Downstream performance of \modelname{} (blue line) by varying the support set size. Dotted lines correspond to the performance of specialist models of each task.}
    \label{fig:performance_on_shots}
    \vspace{-0.4cm}
\end{figure}

\paragraph{Effect of Support Size}
To study the effect of support set size, we plot the performance of \modelname{} with three different shots in Figure~\ref{fig:performance_on_shots}.
We observe that performance consistently increases as support size increases, and beats specialist baselines in all benchmarks with only dozens of labels at most.
This demonstrates the potential of \modelname{} in various dense visual tasks in the wild, whose available supervision ranges between a couple of examples to dozens.

\cutparagraphup
\section{Conclusion}
\cutparagraphup
We proposed \modelname{}, a data-efficient generalist for arbitrary unseen dense visual prediction.
Based on a token-level matching framework, we introduced a flexible encoding mechanism for multiple input images and a powerful task-specific adaptation mechanism for hierarchical architecture.
We have also collected a meta-training dataset by curating six datasets containing diverse dense visual tasks from various domains.
Through extensive experiments, we showed that \modelname{} can learn various unseen tasks with distinct label structures and semantics from training with at most dozens of labels.

\paragraph{Acknowledgements}
This work was supported in part by the National Research Foundation of Korea (RS-2024-00351212), IITP grant (RS-2022-II220926, RS-2022-II220959, and RS-2021-II212068) funded by the Korean government (MSIT), and NAVER-Intel Co-Lab.

\bibliographystyle{splncs04}
\bibliography{main}

\begin{thebibliography}{10}
\providecommand{\url}[1]{\texttt{#1}}
\providecommand{\urlprefix}{URL }
\providecommand{\doi}[1]{https://doi.org/#1}

\bibitem{alayrac2022flamingo}
Alayrac, J.B., Donahue, J., Luc, P., Miech, A., Barr, I., Hasson, Y., Lenc, K.,
  Mensch, A., Millican, K., Reynolds, M., et~al.: Flamingo: a visual language
  model for few-shot learning. Advances in Neural Information Processing
  Systems  \textbf{35},  23716--23736 (2022)

\bibitem{andriluka14cvpr}
Andriluka, M., Pishchulin, L., Gehler, P., Schiele, B.: 2d human pose
  estimation: New benchmark and state of the art analysis. In: IEEE Conference
  on Computer Vision and Pattern Recognition (CVPR) (June 2014)

\bibitem{awais2023foundational}
Awais, M., Naseer, M., Khan, S., Anwer, R.M., Cholakkal, H., Shah, M., Yang,
  M.H., Khan, F.S.: Foundational models defining a new era in vision: A survey
  and outlook. arXiv preprint arXiv:2307.13721  (2023)

\bibitem{bao2022beit}
Bao, H., Dong, L., Piao, S., Wei, F.: {BE}it: {BERT} pre-training of image
  transformers. In: International Conference on Learning Representations
  (2022), \url{https://openreview.net/forum?id=p-BhZSz59o4}

\bibitem{bateni2020improved}
Bateni, P., Goyal, R., Masrani, V., Wood, F., Sigal, L.: Improved few-shot
  visual classification. In: Proceedings of the IEEE/CVF Conference on Computer
  Vision and Pattern Recognition. pp. 14493--14502 (2020)

\bibitem{bay2008speeded}
Bay, H., Ess, A., Tuytelaars, T., Van~Gool, L.: Speeded-up robust features
  (surf). Computer vision and image understanding  \textbf{110}(3),  346--359
  (2008)

\bibitem{brown2020language}
Brown, T., Mann, B., Ryder, N., Subbiah, M., Kaplan, J.D., Dhariwal, P.,
  Neelakantan, A., Shyam, P., Sastry, G., Askell, A., et~al.: Language models
  are few-shot learners. Advances in neural information processing systems
  \textbf{33},  1877--1901 (2020)

\bibitem{caesar2018coco}
Caesar, H., Uijlings, J., Ferrari, V.: Coco-stuff: Thing and stuff classes in
  context. In: Proceedings of the IEEE conference on computer vision and
  pattern recognition. pp. 1209--1218 (2018)

\bibitem{liu2022countr}
Chang, L., Yujie, Z., Andrew, Z., Weidi, X.: Countr: Transformer-based
  generalised visual counting. In: British Machine Vision Conference (BMVC)
  (2022)

\bibitem{chen2017rethinking}
Chen, L.C., Papandreou, G., Schroff, F., Adam, H.: Rethinking atrous
  convolution for semantic image segmentation. arXiv preprint arXiv:1706.05587
  (2017)

\bibitem{chen2023generalist}
Chen, T., Li, L., Saxena, S., Hinton, G., Fleet, D.J.: A generalist framework
  for panoptic segmentation of images and videos. In: Proceedings of the
  IEEE/CVF International Conference on Computer Vision. pp. 909--919 (2023)

\bibitem{cheng2022xmem}
Cheng, H.K., Schwing, A.G.: Xmem: Long-term video object segmentation with an
  atkinson-shiffrin memory model. In: Computer Vision--ECCV 2022: 17th European
  Conference, Tel Aviv, Israel, October 23--27, 2022, Proceedings, Part XXVIII.
  pp. 640--658. Springer (2022)

\bibitem{djukic2023low}
Djukic, N., Lukezic, A., Zavrtanik, V., Kristan, M.: A low-shot object counting
  network with iterative prototype adaptation. In: Proceedings of the IEEE/CVF
  International Conference on Computer Vision. pp. 18872--18881 (2023)

\bibitem{dosovitskiy2021an}
Dosovitskiy, A., Beyer, L., Kolesnikov, A., Weissenborn, D., Zhai, X.,
  Unterthiner, T., Dehghani, M., Minderer, M., Heigold, G., Gelly, S.,
  Uszkoreit, J., Houlsby, N.: An image is worth 16x16 words: Transformers for
  image recognition at scale. In: International Conference on Learning
  Representations (2021), \url{https://openreview.net/forum?id=YicbFdNTTy}

\bibitem{fan2020few}
Fan, Q., Zhuo, W., Tang, C.K., Tai, Y.W.: Few-shot object detection with
  attention-rpn and multi-relation detector. In: Proceedings of the IEEE/CVF
  conference on computer vision and pattern recognition. pp. 4013--4022 (2020)

\bibitem{Fonder2019MidAir}
Fonder, M., Droogenbroeck, M.V.: Mid-air: A multi-modal dataset for extremely
  low altitude drone flights. In: Conference on Computer Vision and Pattern
  Recognition Workshop (CVPRW) (June 2019)

\bibitem{geng2023instructdiffusion}
Geng, Z., Yang, B., Hang, T., Li, C., Gu, S., Zhang, T., Bao, J., Zhang, Z.,
  Hu, H., Chen, D., et~al.: Instructdiffusion: A generalist modeling interface
  for vision tasks. arXiv preprint arXiv:2309.03895  (2023)

\bibitem{han2022few}
Han, G., Ma, J., Huang, S., Chen, L., Chang, S.F.: Few-shot object detection
  with fully cross-transformer. In: Proceedings of the IEEE/CVF conference on
  computer vision and pattern recognition. pp. 5321--5330 (2022)

\bibitem{he2022fully}
He, X., Tan, E.L., Bi, H., Zhang, X., Zhao, S., Lei, B.: Fully transformer
  network for skin lesion analysis. Medical Image Analysis  \textbf{77},
  102357 (2022)

\bibitem{hinterstoisser2011linemod}
Hinterstoisser, S., Holzer, S., Cagniart, C., Ilic, S., Konolige, K., Navab,
  N., Lepetit, V.: Multimodal templates for real-time detection of texture-less
  objects in heavily cluttered scenes. In: 2011 international conference on
  computer vision. pp. 858--865. IEEE (2011)

\bibitem{hong2022cost}
Hong, S., Cho, S., Nam, J., Lin, S., Kim, S.: Cost aggregation with 4d
  convolutional swin transformer for few-shot segmentation. In: European
  Conference on Computer Vision. pp. 108--126. Springer (2022)

\bibitem{ibarz2022generalist}
Ibarz, B., Kurin, V., Papamakarios, G., Nikiforou, K., Bennani, M.,
  Csord{\'a}s, R., Dudzik, A.J., Bo{\v{s}}njak, M., Vitvitskyi, A., Rubanova,
  Y., et~al.: A generalist neural algorithmic learner. In: Learning on Graphs
  Conference. pp.~2--1. PMLR (2022)

\bibitem{kanopoulos1988design}
Kanopoulos, N., Vasanthavada, N., Baker, R.L.: Design of an image edge
  detection filter using the sobel operator. JSSC  (1988)

\bibitem{kim2023universal}
Kim, D., Kim, J., Cho, S., Luo, C., Hong, S.: Universal few-shot learning of
  dense prediction tasks with visual token matching. In: The Eleventh
  International Conference on Learning Representations (2023)

\bibitem{kolesnikov2022uvim}
Kolesnikov, A., Susano~Pinto, A., Beyer, L., Zhai, X., Harmsen, J., Houlsby,
  N.: Uvim: A unified modeling approach for vision with learned guiding codes.
  Advances in Neural Information Processing Systems  \textbf{35},  26295--26308
  (2022)

\bibitem{lepetit2009ep}
Lepetit, V., Moreno-Noguer, F., Fua, P.: Epnp: An accurate o(n) solution to the
  pnp problem. International journal of computer vision  \textbf{81},  155--166
  (2009)

\bibitem{li2023uni}
Li, H., Zhu, J., Jiang, X., Zhu, X., Li, H., Yuan, C., Wang, X., Qiao, Y.,
  Wang, X., Wang, W., et~al.: Uni-perceiver v2: A generalist model for
  large-scale vision and vision-language tasks. In: Proceedings of the IEEE/CVF
  Conference on Computer Vision and Pattern Recognition. pp. 2691--2700 (2023)

\bibitem{li2019cdpn}
Li, Z., Wang, G., Ji, X.: Cdpn: Coordinates-based disentangled pose network for
  real-time rgb-based 6-dof object pose estimation. In: Proceedings of the
  IEEE/CVF International Conference on Computer Vision. pp. 7678--7687 (2019)

\bibitem{lin2017feature}
Lin, T.Y., Doll{\'a}r, P., Girshick, R., He, K., Hariharan, B., Belongie, S.:
  Feature pyramid networks for object detection. In: Proceedings of the IEEE
  conference on computer vision and pattern recognition. pp. 2117--2125 (2017)

\bibitem{lin2014microsoft}
Lin, T.Y., Maire, M., Belongie, S., Hays, J., Perona, P., Ramanan, D.,
  Doll{\'a}r, P., Zitnick, C.L.: Microsoft coco: Common objects in context. In:
  Computer Vision--ECCV 2014: 13th European Conference, Zurich, Switzerland,
  September 6-12, 2014, Proceedings, Part V 13. pp. 740--755. Springer (2014)

\bibitem{liu2020universal}
Liu, L., Hamilton, W.L., Long, G., Jiang, J., Larochelle, H.: A universal
  representation transformer layer for few-shot image classification. In:
  International Conference on Learning Representations (2020)

\bibitem{liuLQWTcvpr16DeepFashion}
Liu, Z., Luo, P., Qiu, S., Wang, X., Tang, X.: Deepfashion: Powering robust
  clothes recognition and retrieval with rich annotations. In: Proceedings of
  IEEE Conference on Computer Vision and Pattern Recognition (CVPR) (June 2016)

\bibitem{lu2023unifiedio}
Lu, J., Clark, C., Zellers, R., Mottaghi, R., Kembhavi, A.: {UNIFIED}-{IO}: A
  unified model for vision, language, and multi-modal tasks. In: The Eleventh
  International Conference on Learning Representations (2023),
  \url{https://openreview.net/forum?id=E01k9048soZ}

\bibitem{mahdavi2022towards}
Mahdavi, S., Swersky, K., Kipf, T., Hashemi, M., Thrampoulidis, C., Liao, R.:
  Towards better out-of-distribution generalization of neural algorithmic
  reasoning tasks. Transactions on Machine Learning Research  (2022)

\bibitem{milton2019isic}
Milton, M.A.A.: Automated skin lesion classification using ensemble of deep
  neural networks in isic 2018: Skin lesion analysis towards melanoma detection
  challenge. arXiv preprint arXiv:1901.10802  (2019)

\bibitem{min2021hypercorrelation}
Min, J., Kang, D., Cho, M.: Hypercorrelation squeeze for few-shot segmentation.
  In: Proceedings of the IEEE/CVF international conference on computer vision.
  pp. 6941--6952 (2021)

\bibitem{openai2023gpt}
OpenAI, R.: Gpt-4 technical report. arxiv 2303.08774. View in Article
  \textbf{2} (2023)

\bibitem{oquab2023dinov2}
Oquab, M., Darcet, T., Moutakanni, T., Vo, H., Szafraniec, M., Khalidov, V.,
  Fernandez, P., Haziza, D., Massa, F., El-Nouby, A., et~al.: Dinov2: Learning
  robust visual features without supervision. arXiv preprint arXiv:2304.07193
  (2023)

\bibitem{ouyang2022training}
Ouyang, L., Wu, J., Jiang, X., Almeida, D., Wainwright, C., Mishkin, P., Zhang,
  C., Agarwal, S., Slama, K., Ray, A., et~al.: Training language models to
  follow instructions with human feedback. Advances in Neural Information
  Processing Systems  \textbf{35},  27730--27744 (2022)

\bibitem{peng2022beit}
Peng, Z., Dong, L., Bao, H., Ye, Q., Wei, F.: Beit v2: Masked image modeling
  with vector-quantized visual tokenizers. arXiv preprint arXiv:2208.06366
  (2022)

\bibitem{jordi2017davis}
Pont-Tuset, J., Perazzi, F., Caelles, S., Arbel\'aez, P., Sorkine-Hornung, A.,
  {Van Gool}, L.: The 2017 davis challenge on video object segmentation.
  arXiv:1704.00675  (2017)

\bibitem{rad2017bb8}
Rad, M., Lepetit, V.: Bb8: A scalable, accurate, robust to partial occlusion
  method for predicting the 3d poses of challenging objects without using
  depth. In: Proceedings of the IEEE international conference on computer
  vision. pp. 3828--3836 (2017)

\bibitem{raffel2020exploring}
Raffel, C., Shazeer, N., Roberts, A., Lee, K., Narang, S., Matena, M., Zhou,
  Y., Li, W., Liu, P.J.: Exploring the limits of transfer learning with a
  unified text-to-text transformer. The Journal of Machine Learning Research
  \textbf{21}(1),  5485--5551 (2020)

\bibitem{ranftl2021vision}
Ranftl, R., Bochkovskiy, A., Koltun, V.: Vision transformers for dense
  prediction. In: Proceedings of the IEEE/CVF International Conference on
  Computer Vision. pp. 12179--12188 (2021)

\bibitem{ranjan2021fsc}
Ranjan, V., Sharma, U., Nguyen, T., Hoai, M.: Learning to count everything. In:
  Proceedings of the IEEE/CVF Conference on Computer Vision and Pattern
  Recognition. pp. 3394--3403 (2021)

\bibitem{reed2022a}
Reed, S., Zolna, K., Parisotto, E., Colmenarejo, S.G., Novikov, A.,
  Barth-maron, G., Gim{\'e}nez, M., Sulsky, Y., Kay, J., Springenberg, J.T.,
  Eccles, T., Bruce, J., Razavi, A., Edwards, A., Heess, N., Chen, Y., Hadsell,
  R., Vinyals, O., Bordbar, M., de~Freitas, N.: A generalist agent.
  Transactions on Machine Learning Research  (2022),
  \url{https://openreview.net/forum?id=1ikK0kHjvj}, featured Certification

\bibitem{rodionov2024neural}
Rodionov, G., Prokhorenkova, L.: Neural algorithmic reasoning without
  intermediate supervision. Advances in Neural Information Processing Systems
  \textbf{36} (2024)

\bibitem{schmidt2018cell}
Schmidt, U., Weigert, M., Broaddus, C., Myers, G.: Cell detection with
  star-convex polygons. In: Medical Image Computing and Computer Assisted
  Intervention--MICCAI 2018: 21st International Conference, Granada, Spain,
  September 16-20, 2018, Proceedings, Part II 11. pp. 265--273. Springer (2018)

\bibitem{schubert2023generalist}
Schubert, I., Zhang, J., Bruce, J., Bechtle, S., Parisotto, E., Riedmiller, M.,
  Springenberg, J.T., Byravan, A., Hasenclever, L., Heess, N.: A generalist
  dynamics model for control. arXiv preprint arXiv:2305.10912  (2023)

\bibitem{shaban2017one}
Shaban, A., Bansal, S., Liu, Z., Essa, I., Boots, B.: One-shot learning for
  semantic segmentation. In: BMVC (2017)

\bibitem{shridhar2023perceiver}
Shridhar, M., Manuelli, L., Fox, D.: Perceiver-actor: A multi-task transformer
  for robotic manipulation. In: Conference on Robot Learning. pp. 785--799.
  PMLR (2023)

\bibitem{snell2017prototypical}
Snell, J., Swersky, K., Zemel, R.: Prototypical networks for few-shot learning.
  Advances in neural information processing systems  \textbf{30} (2017)

\bibitem{steder2010narf}
Steder, B., Rusu, R.B., Konolige, K., Burgard, W.: Narf: 3d range image
  features for object recognition. In: Workshop on Defining and Solving
  Realistic Perception Problems in Personal Robotics at the IEEE/RSJ Int. Conf.
  on Intelligent Robots and Systems (IROS). vol.~44, p.~2. Citeseer (2010)

\bibitem{stringer2021cellpose}
Stringer, C., Wang, T., Michaelos, M., Pachitariu, M.: Cellpose: a generalist
  algorithm for cellular segmentation. Nature methods  \textbf{18}(1),
  100--106 (2021)

\bibitem{sun2019deep}
Sun, K., Xiao, B., Liu, D., Wang, J.: Deep high-resolution representation
  learning for human pose estimation. In: Proceedings of the IEEE/CVF
  conference on computer vision and pattern recognition. pp. 5693--5703 (2019)

\bibitem{valanarasu2022unext}
Valanarasu, J.M.J., Patel, V.M.: Unext: Mlp-based rapid medical image
  segmentation network. In: International Conference on Medical Image Computing
  and Computer-Assisted Intervention. pp. 23--33. Springer (2022)

\bibitem{vaswani2017attention}
Vaswani, A., Shazeer, N., Parmar, N., Uszkoreit, J., Jones, L., Gomez, A.N.,
  Kaiser, {\L}., Polosukhin, I.: Attention is all you need. Advances in neural
  information processing systems  \textbf{30} (2017)

\bibitem{vinyals2016matching}
Vinyals, O., Blundell, C., Lillicrap, T., Wierstra, D., et~al.: Matching
  networks for one shot learning. Advances in neural information processing
  systems  \textbf{29} (2016)

\bibitem{wang2023look}
Wang, J., Chen, D., Wu, Z., Luo, C., Tang, C., Dai, X., Zhao, Y., Xie, Y.,
  Yuan, L., Jiang, Y.G.: Look before you match: Instance understanding matters
  in video object segmentation. In: Proceedings of the IEEE/CVF Conference on
  Computer Vision and Pattern Recognition. pp. 2268--2278 (2023)

\bibitem{wang2020frustratingly}
Wang, X., Huang, T., Gonzalez, J., Darrell, T., Yu, F.: Frustratingly simple
  few-shot object detection. In: International Conference on Machine Learning.
  pp. 9919--9928. PMLR (2020)

\bibitem{wang2022images}
Wang, X., Wang, W., Cao, Y., Shen, C., Huang, T.: Images speak in images: A
  generalist painter for in-context visual learning. arXiv preprint
  arXiv:2212.02499  (2022)

\bibitem{wang2023seggpt}
Wang, X., Zhang, X., Cao, Y., Wang, W., Shen, C., Huang, T.: Seggpt: Towards
  segmenting everything in context. In: Proceedings of the IEEE/CVF
  International Conference on Computer Vision. pp. 1130--1140 (2023)

\bibitem{xiao2018simple}
Xiao, B., Wu, H., Wei, Y.: Simple baselines for human pose estimation and
  tracking. In: Proceedings of the European conference on computer vision
  (ECCV). pp. 466--481 (2018)

\bibitem{ye2022taskprompter}
Ye, H., Xu, D.: Taskprompter: Spatial-channel multi-task prompting for dense
  scene understanding. In: The Eleventh International Conference on Learning
  Representations (2022)

\bibitem{yu2021ap}
Yu, H., Xu, Y., Zhang, J., Zhao, W., Guan, Z., Tao, D.: Ap-10k: A benchmark for
  animal pose estimation in the wild. arXiv preprint arXiv:2108.12617  (2021)

\bibitem{zaken2022bitfit}
Zaken, E.B., Goldberg, Y., Ravfogel, S.: Bitfit: Simple parameter-efficient
  fine-tuning for transformer-based masked language-models. In: Proceedings of
  the 60th Annual Meeting of the Association for Computational Linguistics
  (Volume 2: Short Papers). pp.~1--9 (2022)

\bibitem{zakharov2019dpod}
Zakharov, S., Shugurov, I., Ilic, S.: Dpod: 6d pose object detector and
  refiner. In: Proceedings of the IEEE/CVF international conference on computer
  vision. pp. 1941--1950 (2019)

\bibitem{zamir2018taskonomy}
Zamir, A.R., Sax, A., Shen, W., Guibas, L.J., Malik, J., Savarese, S.:
  Taskonomy: Disentangling task transfer learning. In: Proceedings of the IEEE
  conference on computer vision and pattern recognition. pp. 3712--3722 (2018)

\bibitem{Freihand2019}
Zimmermann, C., Ceylan, D., Yang, J., Russel, B., Argus, M., Brox, T.:
  Freihand: A dataset for markerless capture of hand pose and shape from single
  rgb images. In: IEEE International Conference on Computer Vision (ICCV)
  (2019), \url{"https://lmb.informatik.uni-freiburg.de/projects/freihand/"}

\end{thebibliography}

\clearpage
\appendix
\section*{\LARGE Appendix}

This document provides our implementation details and additional results that could not be accommodated in the main paper due to space limitations.

\setcounter{section}{0}
\renewcommand*{\thesection}{\Alph{section}}

\vspace{-0.3cm}
\section{Meta-Training Details and More Results}
\vspace{-0.2cm}
\label{sec:meta_training_details}

This section describes details of meta-training.
We first describe the dataset details and then describe the implementation details.

\vspace{-0.3cm}
\subsection{Datasets}

We use six existing datasets to construct our unified meta-training dataset.
We summarize the statistics of each dataset in Table~\ref{tab:data_statistics}.

\begin{enumerate}
    \vspace{-0.2cm}
    \item \textbf{Taskonomy}: The Taskonomy dataset~\cite{zamir2018taskonomy} comprises an indoor scene dataset annotated with various vision task labels.
    We utilize a small subset consisting of $\sim$380K images collected from 35 buildings with various camera properties, such as camera pitch, roll, or FoV.
    Following Kim~et~al.~\cite{kim2023universal}, we select 10 dense prediction tasks, including semantic segmentation, surface normal, Euclidean distance, Z-buffer depth, texture edge, occlusion edge, 2d keypoints, 3d keypoints, reshading, and principal curvature.
    Note that the 2d and 3d keypoint labels in the Taskonomy dataset are obtained by descriptor-based algorithms~\cite{bay2008speeded,steder2010narf} and differ from the joint keypoints we describe in the following datasets.
    Additionally, three single-channel tasks (texture edge, occlusion edge, and Euclidean distance) are pre-processed to multi-channel labels.
    Since labels of the texture edge task can be generated by a deterministic edge detection algorithm~\cite{kanopoulos1988design}, we include the unsupervised task in all the following sub-datasets along with two additional tasks (autoencoding and denoising).

    \vspace{0.2cm}
    \item \textbf{COCO}: The COCO dataset~\cite{lin2014microsoft} consists of $\sim$120K images of everyday objects, which contains semantic/instance segmentation annotations of various object categories and human keypoint annotations of 17 keypoint classes.
    With three unsupervised tasks included in the Taskonomy dataset, we include five types of tasks: semantic segmentation, panoptic segmentation, interactive segmentation, joint-specific human keypoint detection, and joint-agnostic human keypoint detection.
    For semantic segmentation and panoptic segmentation, we use 80 object categories from COCO 2017 split and five selected categories (tree, wall-concrete, sky-other, building-other, and grass) from COCO-Stuff~\cite{caesar2018coco} split, respectively.
    For the joint-specific keypoint detection task, we use the human keypoint labels from COCO 2017 split.
    We also add a joint-agnostic keypoint detection task, whose objective is to predict all human joint locations within an image without distinguishing specific ones.
    We categorize this task as continuous signal prediction in Table~\ref{tab:data_statistics}.
    Finally, we include an interactive segmentation task using the instance segmentation labels from COCO 2017 split, which consists of the same 80 object categories used for the semantic segmentation task.
    First, we randomly choose $k \in [1, \lfloor K / 2 \rfloor]$ object instances of a specific class from each image, where $K$ denotes the total number of instances within the image.
    Then the model should segment the chosen instances by using the interactive guide given as a second image, where the guide is generated by a Mixture of Gaussian density whose centers are randomly sampled at $p \in [1, 5]$ pixels from each chosen instance.

    \vspace{0.2cm}
    \item \textbf{MidAir}: The MidAir dataset~\cite{Fonder2019MidAir} consists of $\sim$420K aerial video frames recorded in a synthetic environment.
    It includes two splits (KITE and PLE), each containing videos from multiple trajectories under four distinct weather conditions (sunny, sunset, cloudy, and foggy) and three distinct seasons (spring, fall, winter), respectively.
    For each weather/season condition, we employ the first three trajectories of the KITE split and the first two trajectories of the PLE split as our training data.
    The remaining final trajectory from each split serves as our validation data.
    We select three monocular dense prediction tasks (semantic segmentation, depth estimation, and surface normal) and two binocular dense prediction tasks (stereo depth estimation and stereo surface normal).
    We use images of a left camera for monocular tasks and images of left and right cameras for binocular tasks.
    For semantic segmentation, we use eight categories (trees, dirt ground, ground vegetation, rocky ground, boulders, water plane, road, train track) out of twelve, as the remaining four categories occupy a tiny portion of pixels in the entire dataset.
    We also include three unsupervised tasks as in Taskonomy.

    \vspace{0.2cm}
    \item \textbf{KP-3}: We include three additional datasets consisting of joint keypoint labels.
    \textbf{MPII}~\cite{andriluka14cvpr} dataset consists of $\sim$25K human images annotated with 16 human joints, whose joint definition differs from the COCO dataset. \textbf{DeepFashion}~\cite{liuLQWTcvpr16DeepFashion} dataset consists of $\sim$120K fashion images annotated with 8 fashion landmarks as joints. Finally, \textbf{FreiHand}~\cite{Freihand2019} dataset consists of $\sim$130K hand images captured from 32 subjects, annotated with 21 hand joints.
    Similar to the COCO dataset, we both include joint-specific and joint-agnostic keypoint detection tasks in all of the three keypoint-specific datasets, as well as the unsupervised tasks.
    In Table~\ref{tab:data_ablation_table}, we refer to this dataset together with the COCO keypoint dataset as KP-4.
\end{enumerate}

\begin{table*}[ht!]
    \caption{Statistics of six datasets contained in the meta-training dataset. When counting the number of tasks, we consider different channels of a multi-channel task as different tasks. For example, the number of segmentation tasks corresponds to the number of classes, and the number of joint-specific keypoint tasks corresponds to the number of joints. Numbers in parentheses denote the number of task groups obtained by considering different channels of a multi-channel task and tasks from different source datasets as a single group.}
    \label{tab:data_statistics}
    \vspace{-0.4cm}
    \begin{center}
        \renewcommand{\arraystretch}{1.3}
        \renewcommand{\aboverulesep}{0pt}
        \renewcommand{\belowrulesep}{0pt}
        \setlength\tabcolsep{2pt}
        \footnotesize
        \begin{tabular}{c p{0.01cm} c p{0.01cm} ccc p{0.01cm} cc}
            \toprule
            \multirow{3}{*}{Dataset} & & \multirow{3}{*}{\# Images} & & \multicolumn{6}{c}{\# Tasks} \\
            & & & &
            \multicolumn{3}{c}{Segmentation} & & \multicolumn{2}{c}{Continuous Signals} \\
            & & & &
            semantic & panoptic & interatctive & &
            monocular & binocular
            \\
            
            \midrule
            Taskonomy & &
            381,916 & &
            12 (1) & 0 & 0 & &
            19 (8) & 0
            \\

            COCO & &
            118,287 & &
            80 (1) & 5 (1) & 80 (1) & &
            1 (1) & 0
            \\

            MidAir & &
            423,676 & &
            8 (1) & 0 & 0 & &
            4 (2) & 4 (2)
            \\

            MPII & &
            24,984 & &
            0 & 0 & 0 & &
            1 (1) & 0
            \\

            DeepFashion & &
            123,016 & &
            0 & 0 & 0 & &
            1 (1) & 0
            \\

            FreiHand & &
            130,240 & &
            0 & 0 & 0 & &
            1 (1) & 0
            \\

            \midrule

            Total & &
            1,202,119 & &
            100 (1) & 5 (1) & 80 (1) & &
            27 (8) & 4 (2)
            \\
            
            \bottomrule
        \end{tabular}
        
        \vspace{0.2cm}
        \begin{tabular}{c p{0.01cm} c p{0.01cm} ccc p{0.01cm} c}
            \toprule
            \multirow{3}{*}{Dataset} & & \multicolumn{7}{c}{\# Tasks} \\
            & &
            Keypoints & & \multicolumn{3}{c}{Low-Level} & & \multirow{2}{*}{Total} \\
            & &
            joint-specific & &
            autoencoding & denoising & texture edge & &
            \\
            
            \midrule
            Taskonomy & &
            0 & & 
            3 (1) & 3 (1) & 3 (1) & &
            40 (12)
            \\

            COCO & &
            17 (1) & & 
            3 (1) & 3 (1) & 3 (1) & &
            192 (8)
            \\

            MidAir & &
            0 & & 
            3 (1) & 3 (1) & 3 (1) & &
            25 (8)
            \\

            MPII & &
            16 (1) & & 
            3 (1) & 3 (1) & 3 (1) & &
            26 (5)
            \\

            DeepFashion & &
            8 (1) & & 
            3 (1) & 3 (1) & 3 (1) & &
            18 (5)
            \\

            FreiHand & &
            21 (1) & & 
            3 (1) & 3 (1) & 3 (1) & &
            31 (5)
            \\

            \midrule

            Total & &
            62 (1) & & 
            18 (1) & 18 (1) & 18 (1) & &
            332 (17) \\
            
            \bottomrule
        \end{tabular}
        \vspace{-0.4cm}
    \end{center}
\end{table*}

\subsection{Implementation Details}

\paragraph{Task Sampling}
We train \modelname{} with 400,000 episodic training iterations.
At each iteration of meta-training, we construct 16 episodic batches, where 8 of them consist of a uni-modal task and the remaining consist of a bi-modal task sampled from the unified dataset.
Then we use either batch for computing the loss at each iteration, where the batch is randomly chosen with probability proportional to the number of uni-modal and bi-modal tasks ($246 : 84$).
Both episodic batches are sampled as the following procedure.
\begin{enumerate}
    \vspace{-0.2cm}
    \item First, we sample the type of tasks, which is one of categorical (both segmentation and joint-specific keypoints), continuous, and low-level, with a sampling rate $3 : 3 : 1$.

    \item Second, we sample the source dataset, where the sampling rate is proportional to either the number of tasks (for categorical and continuous tasks) or the number of images (for low-level tasks) included in each dataset.

    \item Third, we uniformly sample four tasks from the task pool filtered by the chosen task type and source dataset, where multi-channel tasks are disassembled to separate single-channel tasks.

    \item Finally, we sample four support pairs and four query pairs from the source dataset with the selected task, thus simulating a $4$-shot learning episode with four queries for each task.
\end{enumerate}

\paragraph{Loss Function}
We use three different loss functions for meta-training: L1 loss, binary cross-entropy (bce) loss, and spatial softmax loss.
We use L1 loss as default while using bce loss for segmentation tasks and spatial softmax loss for joint-specific keypoint detection tasks.
Spatial softmax loss is computed by first applying the softmax function on the prediction along the spatial axis (both horizontal and vertical), then applying the binary cross-entropy loss.
We normalize the spatial softmax loss by the sum of the target label and do not use any other hyper-parameters for weighting three loss functions.

\begin{figure}[t!]
    \centering
    \includegraphics[width=0.9\linewidth]{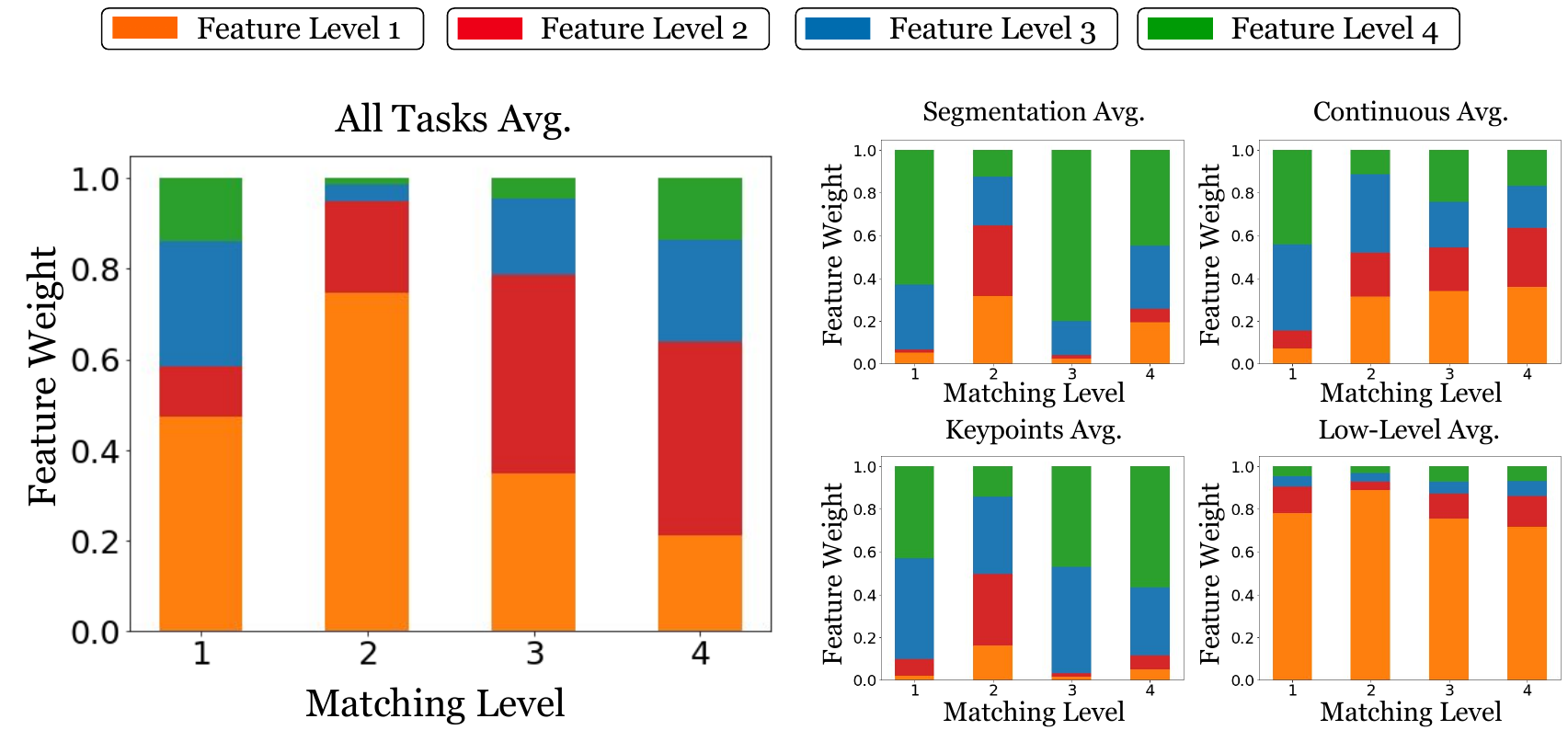}
    \vspace{-0.4cm}
    \caption{Learned feature weights in training tasks.}
    \label{fig:feature_weights_training}
    \vspace{-0.4cm}
\end{figure}

\vspace{0.1cm}
\paragraph{Task-Specific Parameters}
Following VTM~\cite{kim2023universal}, we use task-specific bias parameters of the image encoder for each task, which results in a total of 332 bias sets for meta-training.
In Section~\ref{subsec:multi_input} and Section~\ref{subsec:feature_reweighting}, we have introduced the position bias and the feature re-weighting matrix, which are also task-specific.
During meta-training, we share the position bias for all tasks and fine-tune it task-specifically for downstream tasks with multi-modal inputs.
We share the feature weighting matrix among tasks that originate from the same multi-channel task (\emph{e.g.,} three channels of surface normal) since they would require similar correspondence between image and label features.
We also use the shared matrix for semantic segmentation and panoptic segmentation in COCO.
This results in 42 matrices of size $4 \times 4$, where we use $L=4$ feature levels.
In Figure~\ref{fig:feature_weights_training}, we plot the learned feature weights averaged on all tasks and tasks within each task category in the meta-training dataset.
It clearly shows that different task groups require different feature correspondences, where low-level tasks mainly require low-level features while the other tasks require both low-level and high-level features.

\vspace{0.1cm}
\paragraph{Implementation Framework}
We implemented our model based on PyTorch Lightning which supports both Intel Gaudi-v2 (HABANA) and NVIDIA AI accelerators (CUDA). 
We provide the code for both systems on separate branches in the GitHub repository.

\section{Downstream Details and More Results}
\vspace{-0.2cm}
\label{sec:downstream_details}

In this section, we describe the detailed settings of six downstream tasks and provide additional qualitative results for each task.

\vspace{-0.3cm}
\subsection{AP-10K}

\vspace{-0.1cm}
\paragraph{Detailed Settings}
We fine-tune and evaluate our model on the AP-10K train and test split, respectively.
Since the ground-truth bounding box labels are provided in the dataset, we use it to center-crop the images and labels, then resize them to $256 \times 256$ resolution.
During fine-tuning, we apply a random crop of crop size $224 \times 224$ on the resized data.
During inference, we further resize the data to $224 \times 224$ to obtain the prediction, then translate the keypoint locations back to the original resolution for evaluation.
We use spatial softmax loss described in Section~\ref{sec:meta_training_details}.

\begin{figure}[!t]
    \centering
    \includegraphics[width=0.96\linewidth]{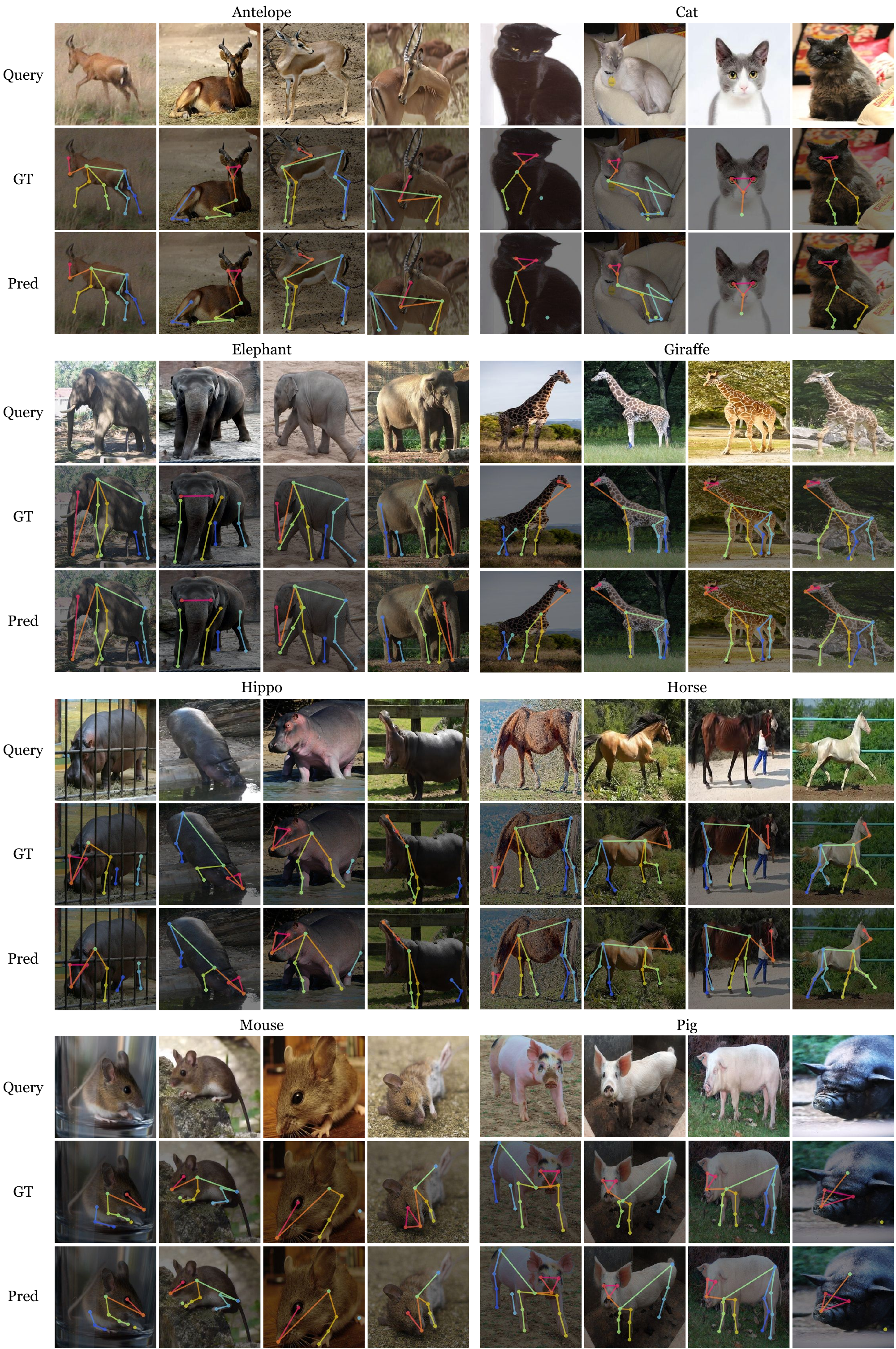}
    \caption{Additional Qualitative Results on AP-10K.}
    \label{fig:ap10k_supp}
\end{figure}

\vspace{0.1cm}
\paragraph{Additional Results}
In Figure~\ref{fig:ap10k_supp}, we provide additional qualitative results on AP-10K.
We can observe that \modelname{} successfully adapts to different animal species with distinctive appearances and joint configurations.

\begin{figure*}[!t]
    \centering
    \includegraphics[width=0.98\linewidth]{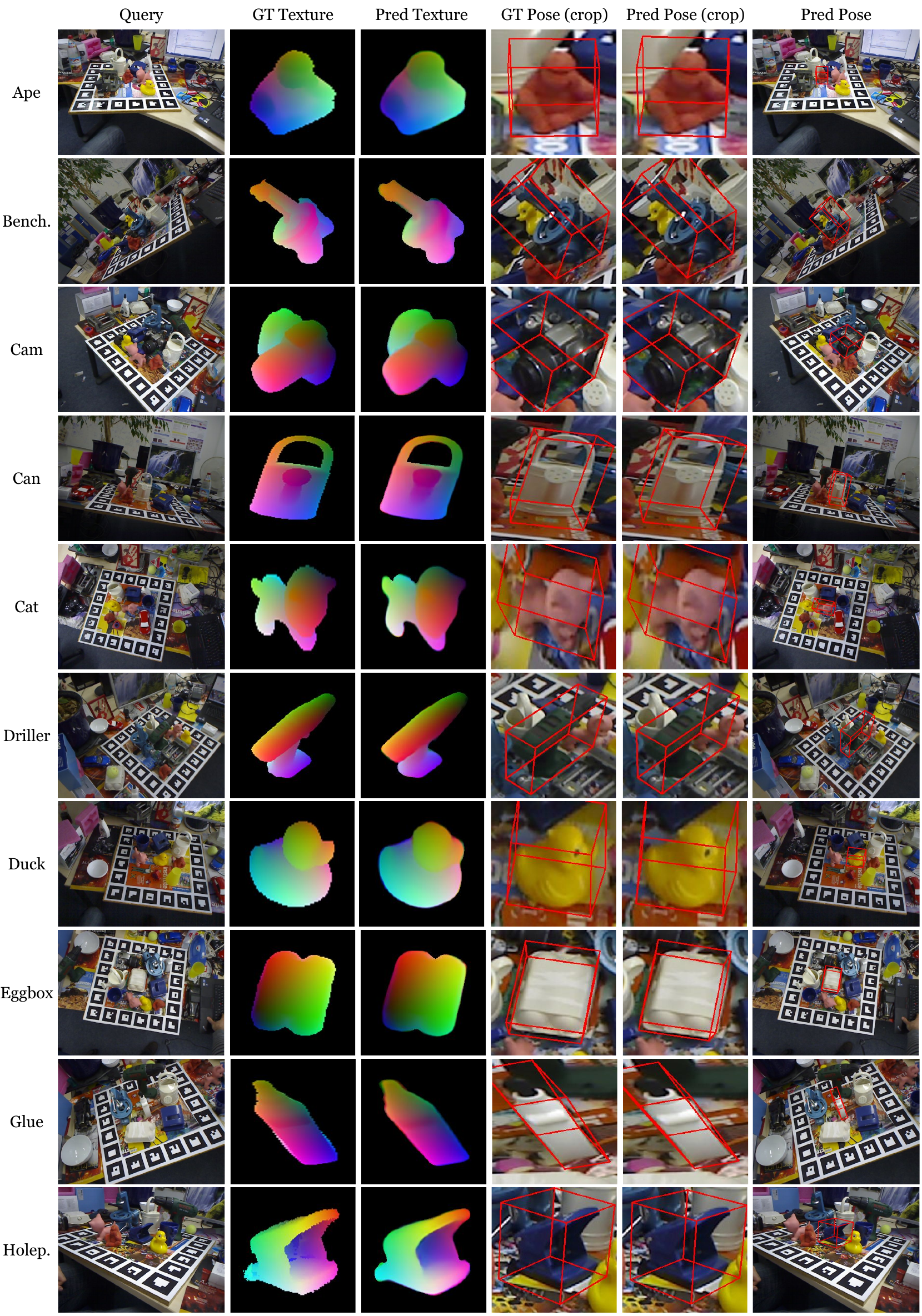}
    \caption{Additional Qualitative Results on LineMOD on ten object classes: Ape, Benchviseblue, Cam, Can, Cat, Driller, Duck, Eggbox, Glue, and Holepuncher.}
    \label{fig:linemod_supp_1}
\end{figure*}

\vspace{-0.2cm}
\subsection{LineMOD}
\vspace{-0.1cm}

\paragraph{Detailed Settings}
We fine-tune and evaluate our model on the conventional train and test split of the LineMOD dataset following literature~\cite{zakharov2019dpod,li2019cdpn}.
As discussed in Section~\ref{sec:experiments}, we formulate a 6D pose estimation task as dense prediction by predicting correspondence between each image pixel and the vertex of the CAD model, from which 6D pose is obtained by Perspective-n-point algorithm~\cite{lepetit2009ep}.
To predict the 2D-3D correspondence, we render a 3-channel texture map on all images using the 6D extrinsic camera matrices, where the channels correspond to the X, Y, and Z axes of the relative 3D position of CAD vertices which are normalized to $[-1, 1]$.
Then we use the rendered texture maps as dense labels to be predicted by \modelname{}.
Since the 2D-3D correspondence is defined on the object area, we augment the dense label with a foreground segmentation channel, which represents a non-zero region of the texture maps, and let \modelname{} also predict it.
We use bce loss to train the segmentation channel and L1 loss to train the texture channels.
During fine-tuning, we apply random rotation, random jittering, and random Gaussian blur as data augmentation as well as random cropping with crop size $224 \times 224$.

\vspace{0.1cm}
\paragraph{Image Cropping with Object Segmentation}
Since the objects occupy a small region of the full image, \modelname{} first predicts the object location to crop the images and then predicts the 6D pose in the cropped region as described above.
To obtain the object location, we perform an additional fine-tuning stage for foreground segmentation on the full-sized images before predicting texture maps.
Then we get the bounding box of the object from the predicted segmentation mask.
To fine-tune the object segmentation, we resize the full images and labels to $256 \times 256$ and apply random cropping with crop size $224 \times 224$, with data augmentation applied in the texture fine-tuning stage.
Note that we generate the segmentation labels from the texture map labels contained in the support set (\emph{i.e.}, non-zero region of the texture maps), \modelname{} does not use additional supervision for the object detection procedure.
After obtaining the bounding box, we center-crop the images and labels and resize them to $256 \times 256$.

\begin{figure*}[!t]
    \centering
    \includegraphics[width=\linewidth]{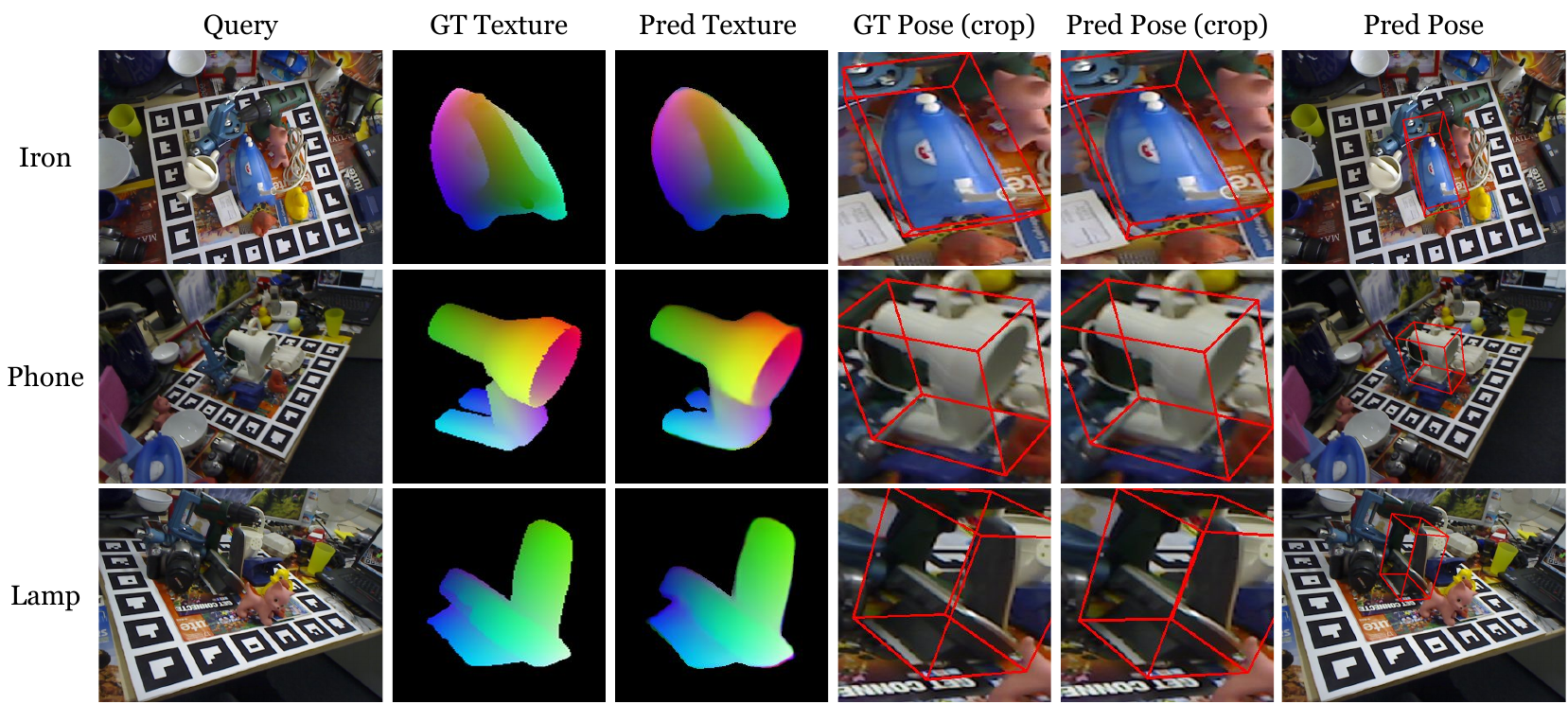}
    \caption{Additional Qualitative Results on LineMOD on three object classes: Iron, Lamp, and Phone.}
    \label{fig:linemod_supp_2}
\end{figure*}

\vspace{0.2cm}
\paragraph{Additional Results}
In Figure~\ref{fig:linemod_supp_1} and Figure~\ref{fig:linemod_supp_2}, we provide additional qualitative results on LineMOD.
We visualize the original query image, texture maps, and 6D pose on the cropped region (both ground truth and prediction), and predicted 6D pose on the original image.
We observe that \modelname{} accurately predicts various 6D poses of all objects.

\subsection{ISIC 2018}

\paragraph{Detailed Settings}
We follow the standard protocol of the literature~\cite{he2022fully,valanarasu2022unext} for fine-tuning and evaluation: (1) we perform 5-fold cross-validation on ISIC 2018 train split and report the mean F1 score, and (2) resize the data to $512 \times 512$ resolution for evaluation.
During fine-tuning, we resize the data to $448 \times 448$ resolution and apply random cropping with crop size $384 \times 384$.
During inference, we first resize the data to $384 \times 384$ to obtain the prediction, then upsample the prediction to $512 \times 512$ resolution for evaluation.
We use bce loss for fine-tuning.

\begin{figure*}[!ht]
    \centering
    \includegraphics[width=\linewidth]{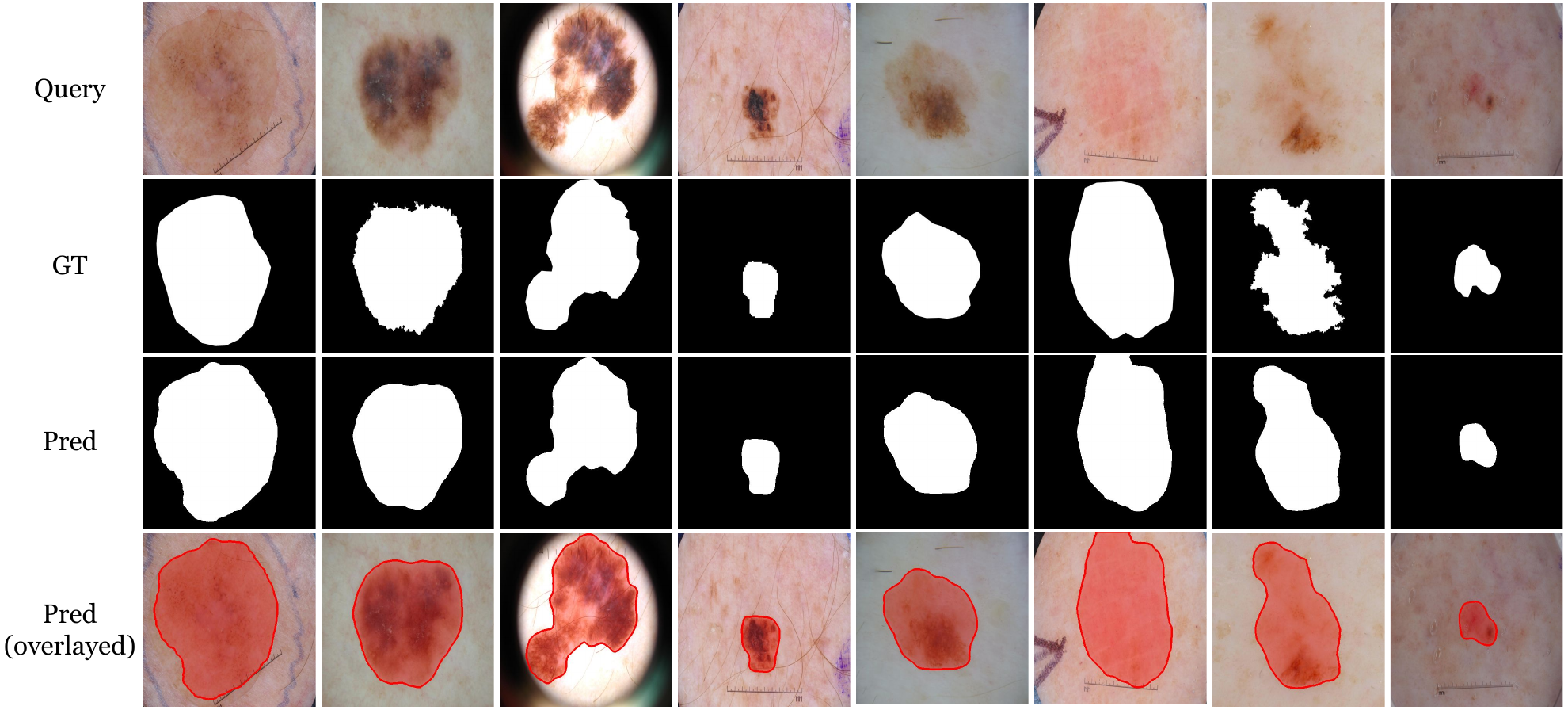}
    \caption{Additional Qualitative Results on ISIC 2018.}
    \label{fig:isic_supp}
\end{figure*}

\begin{figure*}[!ht]
    \centering
    \includegraphics[width=\linewidth]{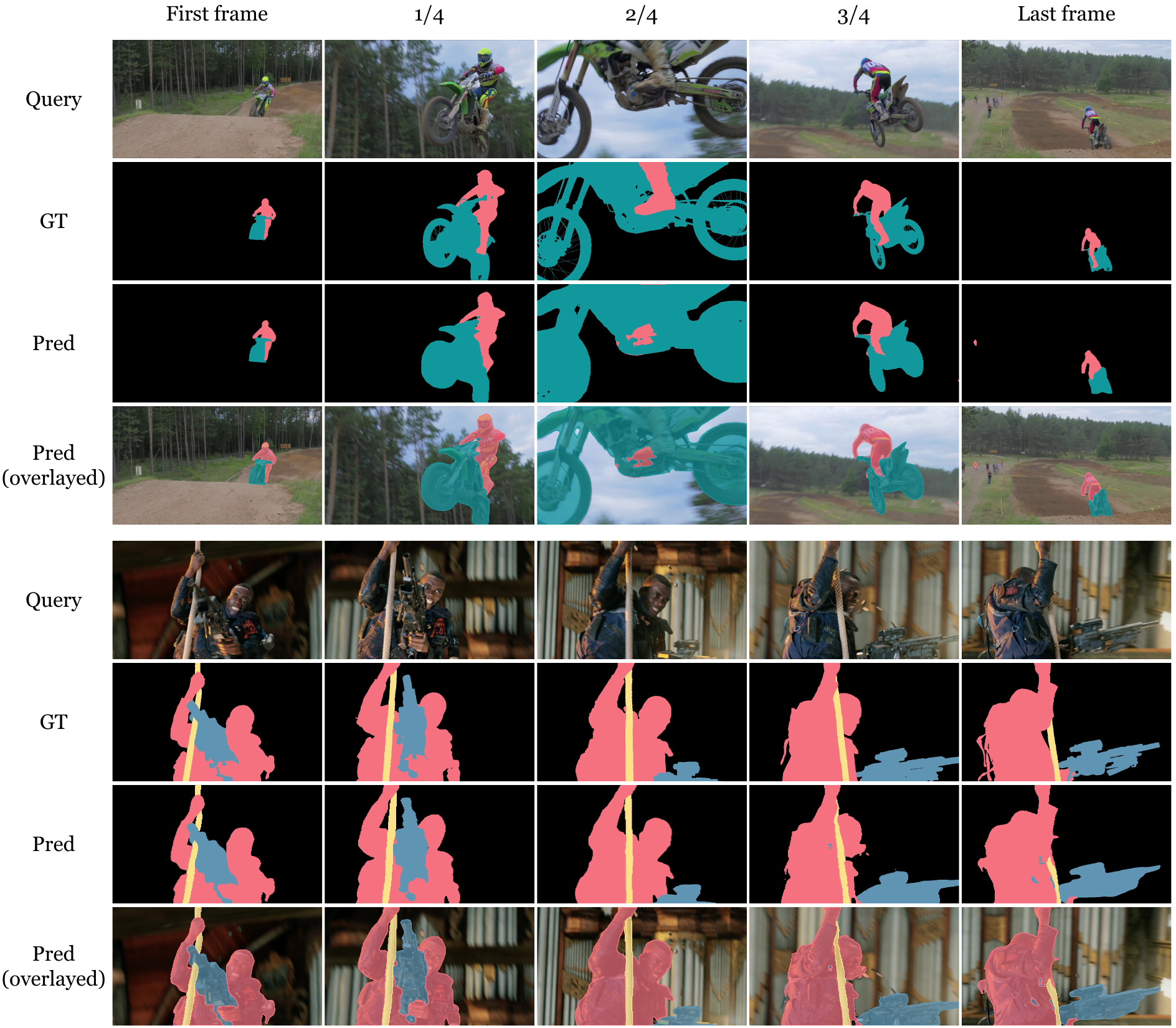}
    \caption{Additional Qualitative Results on DAVIS 2017: motocross-jump and shooting.}
    \label{fig:davis_supp1}
\end{figure*}

\begin{figure*}[!ht]
    \centering
    \vspace{1cm}
    \includegraphics[width=\linewidth]{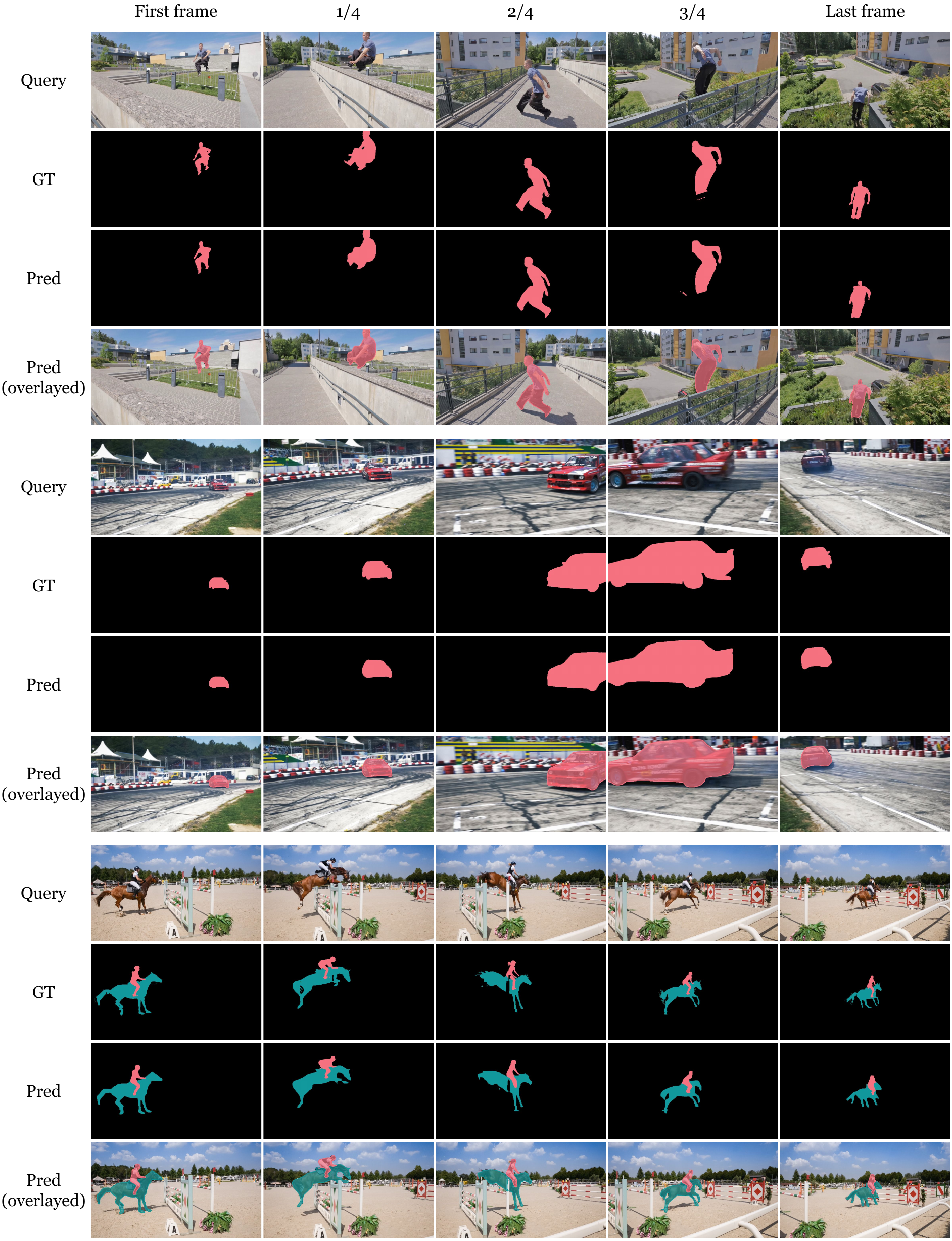}
    \caption{Additional Qualitative Results on DAVIS 2017: parkour, drift-straight, and horse-jump.}
    \label{fig:davis_supp2}
\end{figure*}

\paragraph{Additional Results}
In Figure~\ref{fig:isic_supp}, we provide additional qualitative results on ISIC 2018.
We observe that \modelname{} accurately segments the lesion boundary, even for ambiguous regions like the first and sixth query examples in the figure.

\subsection{DAVIS 2017}

\paragraph{Detailed Settings}
We fine-tune and evaluate our model on the DAVIS 2017 validation split which consists of 30 videos.
Notably, we do not use the videos in the train split of DAVIS 2017, unlike the video object segmentation literature~\cite{cheng2022xmem,wang2023look}.
During fine-tuning, we resize the video frames and labels to $448 \times 448$, then apply random cropping with crop size $384 \times 384$.
During inference, we first resize the data to $384 \times 384$ to obtain the prediction, then upsample the prediction to the original resolution for evaluation.
We treat different instances as different tasks (\emph{e.g.,} we fine-tune five independent task-specific parameters for videos containing five instances).
We use cross-entropy loss over predictions on all object instances within the video, where the logits for the background are fixed to zero when computing the loss.

\vspace{0.2cm}
\paragraph{Additional Results}
In Figure~\ref{fig:davis_supp1} and Figure~\ref{fig:davis_supp2}, we provide additional qualitative results on DAVIS 2017.
We can observe that \modelname{} accurately segments diverse videos within the benchmark.
As discussed in Section~\ref{sec:experiments}, \modelname{} is robust to the variations in object appearance and the camera view, without using any temporal prior.

\begin{figure*}[t!]
    \centering
    \includegraphics[width=\linewidth]{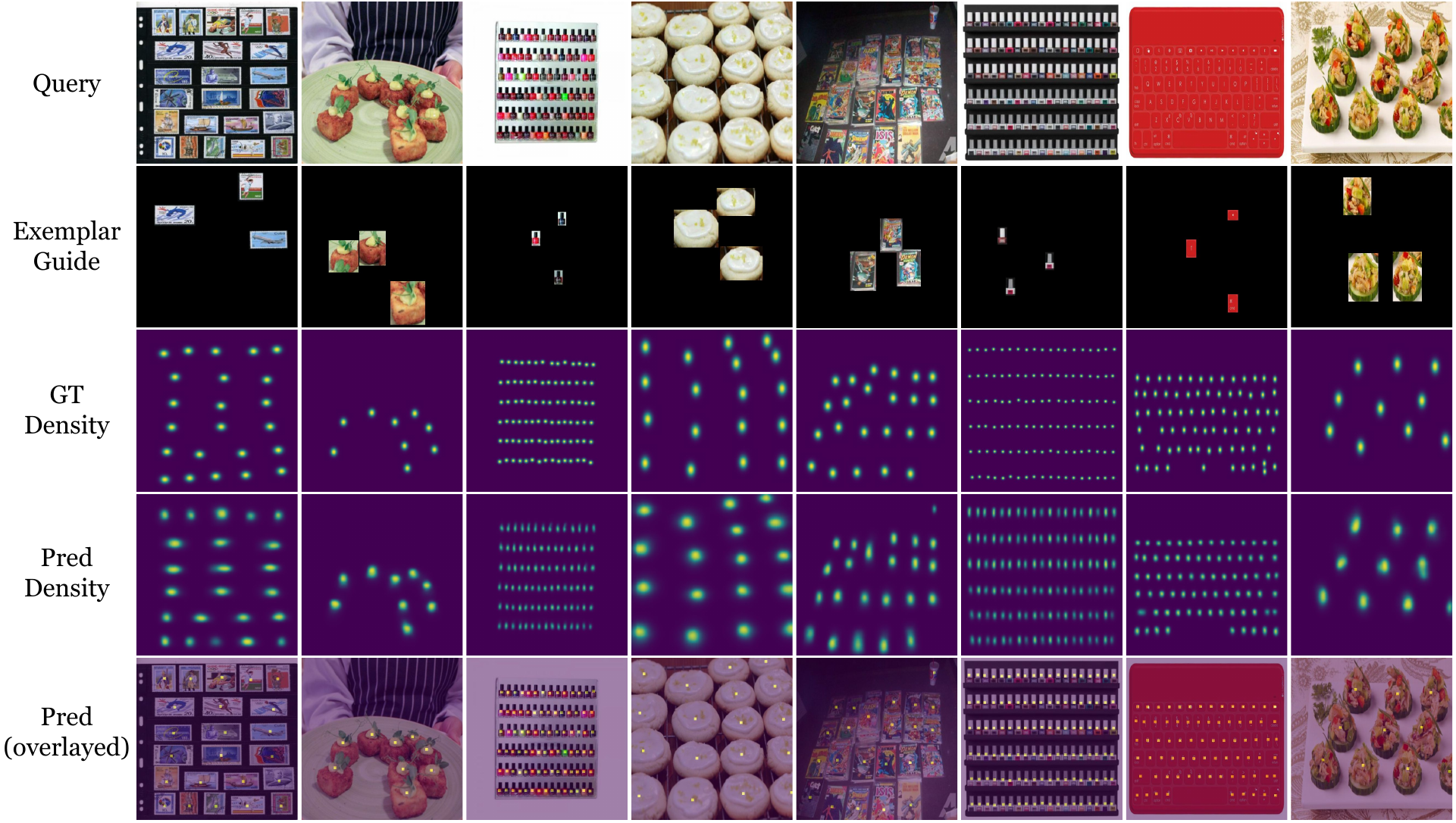}
    \caption{Additional Qualitative Results on FSC-147.}
    \vspace{-0.4cm}
    \label{fig:fsc_supp}
\end{figure*}

\subsection{FSC-147}

\paragraph{Detailed Settings}
We fine-tune and evaluate our model in the train and test split of the FSC-147 dataset, respectively.
We convert the object counting to Gaussian density prediction following the literature~\cite{liu2022countr,djukic2023low}.
After predicting the density map, we detect the modes and use the number of modes as count prediction.
As an exception, for outliers whose sum of the predicted density map is more than 3,000, we use the sum of the density map as count prediction.
To generate the exemplar guide, we copy and paste the exemplar patches to a black image using their bounding boxes, as shown in Figure~\ref{fig:fsc_supp}
We randomly scale the patch size and paste each patch multiple times to maximize the augmentation effect.
For fine-tuning, we first resize the data to $592 \times 592$ and apply random cropping of crop size $512 \times 512$ during fine-tuning.
During inference, we process the images differently depending on the average size of exemplar patches following \cite{liu2022countr}.
For images whose average size of exemplar patches is less than the threshold (13.33 pixels for image size $512 \times 512$), we first resize the data to $1536 \times 1536$ and crop the image into 9 non-overlapping patches of size $512 \times 512$, obtain the predictions separately, then merge the predictions.
For the other images, we resize the data to $512 \times 512$.
We use MSE loss for fine-tuning.

\paragraph{Additional Results}
In Figure~\ref{fig:fsc_supp}, we provide additional qualitative results on FSC-147.
We observe that \modelname{} accurately predicts the density map on the query objects by effectively exploiting the exemplar guide.

\begin{figure*}[t!]
    \centering
    \includegraphics[width=\linewidth]{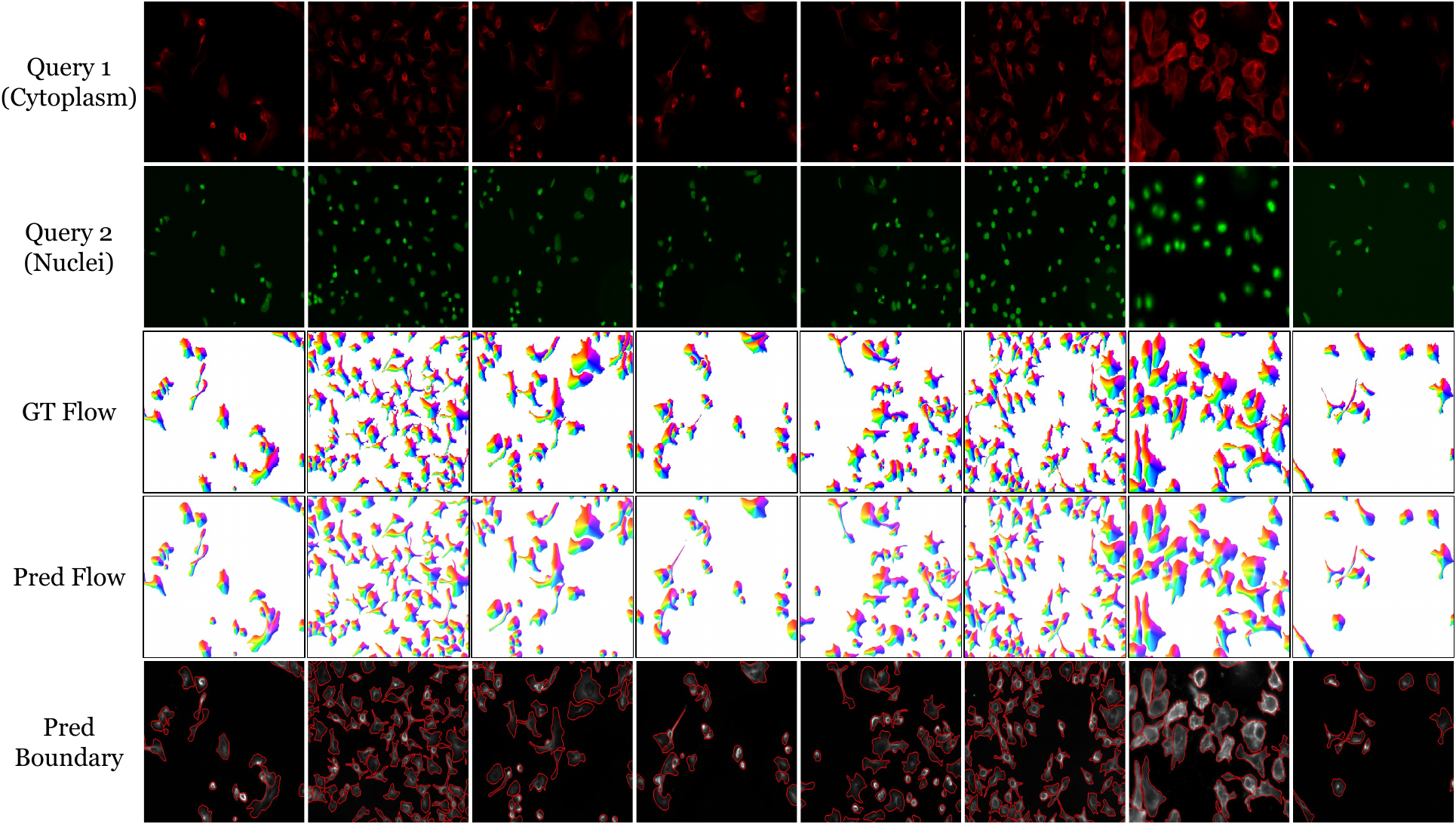}
    \caption{Additional Qualitative Results on Cellpose for bi-modal images. We use red and green color codings for the cytoplasm and nuclie images, respectively.}
    \label{fig:cellpose_supp_1}
\end{figure*}

\subsection{Cellpose}

\paragraph{Detailed Settings}
We fine-tune and evaluate our model in the train and test split of the Cellpose dataset, respectively.
Following \cite{stringer2021cellpose}, we formulate a cell instance segmentation task by flow estimation with foreground mask segmentation.
We generate a 2-channel flow map where each channel corresponds to vertical and horizontal gradients of each cell towards its center.
Then \modelname{} predicts the flow map and also a binary segmentation mask to segment all foreground cells, from which we can obtain the instance segmentation mask.
We resize the images and labels to $256 \times 256$ and apply random resized cropping of crop size $224 \times 224$ and scale between 0.75 and 1.25 during fine-tuning.
During inference, we first divide an image into overlapping tiles with the size of $224 \times 224$ and 50\% overlap, then ensemble each prediction by multiplying it with a Gaussian kernel to minimize edge effects.
For each tile, we additionally use test-time augmentation where each prediction is obtained by the ensemble of 4 flipped inputs following \cite{stringer2021cellpose}.
We use bce loss for the segmentation channel and L1 loss for the flow channels.
We apply random flipping as data augmentation together with random resized cropping.

\begin{figure*}[t!]
    \centering
    \includegraphics[width=\linewidth]{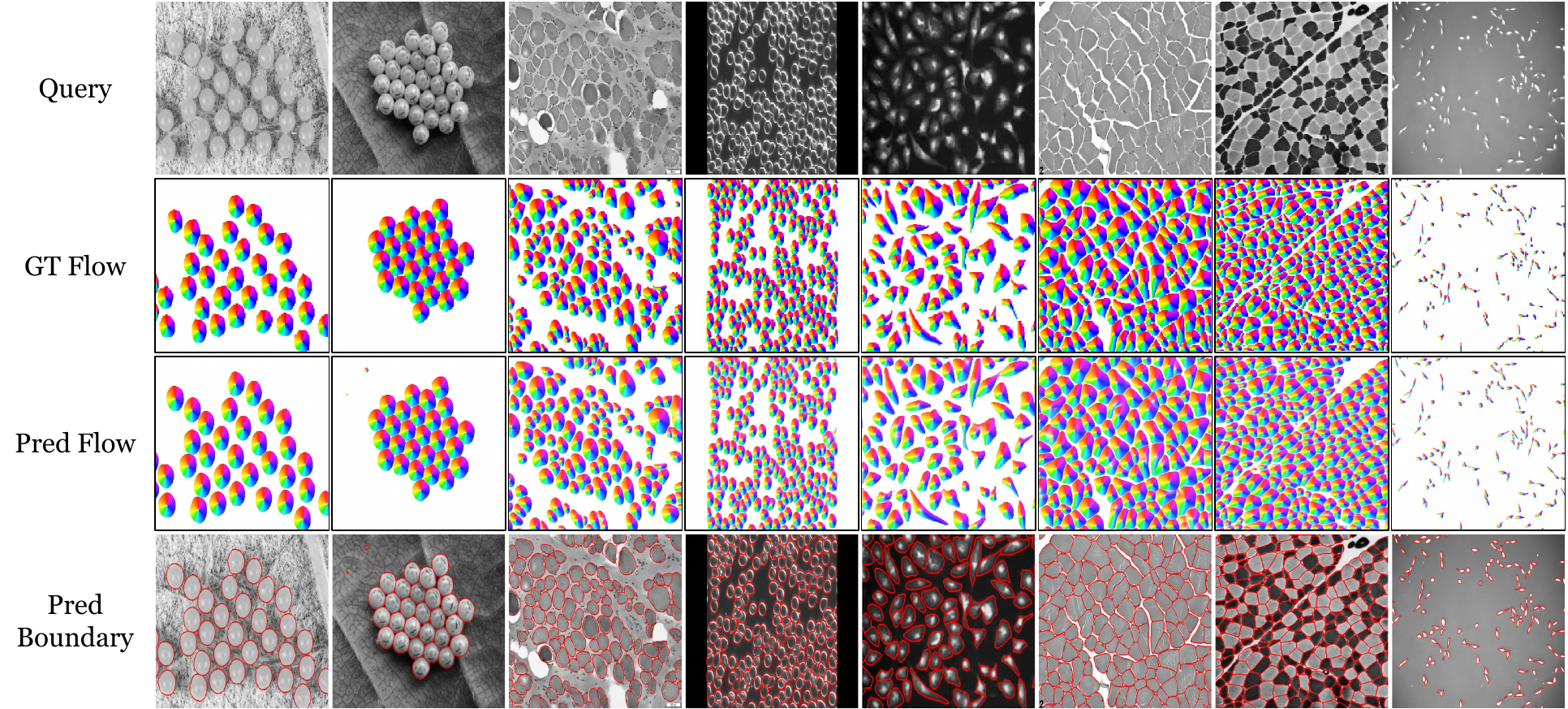}
    \caption{Additional Qualitative Results on Cellpose for uni-modal images. We use grayscale color codings for the cytoplasm images.}
    \label{fig:cellpose_supp_2}
\end{figure*}

\vspace{0.2cm}
\paragraph{Additional Results}
In Figure~\ref{fig:cellpose_supp_1} and Figure~\ref{fig:cellpose_supp_2}, we provide additional qualitative results on Cellpose.
We observe that \modelname{} accurately predicts the cell boundaries for both bi-modal and uni-modal images.

\section{Additional Experiments}
\vspace{-0.2cm}
In this section, we conduct ablation studies on the image encoder backbone and image resolution to analyze their effect on downstream performance.

\vspace{-0.3cm}
\subsection{Ablation Study on Image Encoder Backbone}
\vspace{-0.1cm}
Since the image encoder plays a central role in the matching architecture of \modelname{}, it is important to leverage a pre-trained image encoder backbone as initialization for meta-training.
To analyze the effect of the image encoder backbone, we compare three different pre-trained transformers: BEiTv2 Large~\cite{peng2022beit} (default setting), ViT Large~\cite{dosovitskiy2021an}, and DINOv2 Large~\cite{oquab2023dinov2}.
BEiTv2 and DINOv2 are pre-trained with self-supervised learning objectives, while ViT is pre-trained with image classification.
Also, DINOv2 Large is distilled from the DINOv2 Giant backbone, which is trained on a large-scale dataset containing ImageNet-22k.
In Table~\ref{tab:backbone_ablation_table}, we report the performance of \modelname{} with three different backbones.
We note that BEiTv2 achieves the best performance, while ViT shows inferior performance compared to the self-supervised learning approaches.
This can be attributed to the generality of self-supervised features compared to image classification features which may not have fine-grained information for dense prediction.

\begin{table}[!t]
\caption{Ablation study on the image encoder backbone.}
\label{tab:backbone_ablation_table}
\begin{center}
    \renewcommand{\arraystretch}{1.2}
    \renewcommand{\aboverulesep}{0pt}
    \renewcommand{\belowrulesep}{0pt}
    \setlength\tabcolsep{2pt}
    \small
    \begin{tabular}{c | c p{0.01cm} c p{0.01cm} c p{0.01cm} c p{0.001cm} c p{0.001cm} c}
        \toprule
        \multirow{2}{*}{Backbone}
        
        &
        AP-10K & &
        LineMOD & &
        ISIC 2018 & &
        DAVIS 2017 & &
        FSC-147 & &
        Cellpose
        \\

        \cmidrule{2-12}
        
        &
        AP ↑ & &
        ADD ↑ & &
        F1 ↑ & &
        $\mathcal{J}$\&$\mathcal{F}$ ↑ & &
        MAE ↓ & &
        $\text{AP}_{50}$ ↑
        \\

        \midrule

        ViT~\cite{dosovitskiy2021an} &
        36.3 & &
        66.7 & &
        85.5 & &
        64.9 & &
        26.5 & &
        59.9
        \\

        DINOv2~\cite{oquab2023dinov2} &
        63.5 & &
        85.1 & &
        87.5 & &
        70.4 & &
        20.1 & &
        69.0
        \\

        BEiTv2~\cite{peng2022beit} &
        \textbf{67.2} & &
        \textbf{85.2} & &
        \textbf{88.5} & &
        \textbf{77.5} & &
        \textbf{12.3} & &
        \textbf{70.3}
        \\

        \bottomrule
    \end{tabular}
\end{center}

\end{table}

\vspace{-0.3cm}
\subsection{Ablation Study on Image Resolution}
\vspace{-0.1cm}
We also conduct an ablation study on the image resolution.
Since we use a larger resolution for three downstream benchmarks (ISIC 2018, DAVIS 2017, and FSC-147) compared to the meta-training ($224 \times 224$), we analyze the effect of increasing the resolution.
We observe that using larger image resolution improves the performance at DAVIS 2017 and FSC-147 to a great extent while having a statistically non-significant effect on ISIC 2018.

\begin{table}[!t]
\caption{Ablation study on the image resolution. Larger resolution corresponds to input image size $384 \times 384$ for ISIC 2018 and DAVIS 2017, and $512 \times 512$ for FSC-147.}
\label{tab:resolution_ablation_table}
\begin{center}
    \renewcommand{\arraystretch}{1.2}
    \renewcommand{\aboverulesep}{0pt}
    \renewcommand{\belowrulesep}{0pt}
    \setlength\tabcolsep{2pt}
    \small
    \begin{tabular}{c | c p{0.01cm} c p{0.01cm} c}
        \toprule
        \multirow{2}{*}{Resolution} &
        
        ISIC 2018 & &
        DAVIS 2017 & &
        FSC-147
        \\

        \cmidrule{2-6}
        
        &
        F1 ↑ & &
        $\mathcal{J}$\&$\mathcal{F}$ ↑ & &
        MAE ↓
        \\

        \midrule

        $224 \times 224$ &
        \textbf{88.6} $\pm$ 1.20 & &
        65.7 & &
        32.2
        \\

        Larger Resolution &
        88.5 $\pm$ 0.75 & &
        \textbf{77.5} & &
        \textbf{12.0}
        \\

        \bottomrule
    \end{tabular}
\end{center}
\end{table}

\end{document}